\pdfoutput=1
\documentclass[10pt, a4paper]{article}
\usepackage{lrec2022} 
\usepackage{multibib}
\usepackage{graphicx}
\usepackage{tabularx}
\usepackage{longtable}
\usepackage{soul}
\usepackage{titlesec}
\titleformat{\section}{\normalfont\large\bfseries\center}{\thesection.}{1em}{}
\titleformat{\subsection}{\normalfont\SmallTitleFont\bfseries\raggedright}{\thesubsection.}{1em}{}
\titleformat{\subsubsection}{\normalfont\normalsize\bfseries\raggedright}{\thesubsubsection.}{1em}{}
\renewcommand\thesection{\arabic{section}}
\renewcommand\thesubsection{\thesection.\arabic{subsection}}
\renewcommand\thesubsubsection{\thesubsection.\arabic{subsubsection}}

\usepackage{epstopdf}
\usepackage[utf8]{inputenc}
\usepackage[T1]{fontenc}

\usepackage{hyperref}
\hypersetup{breaklinks=true, colorlinks=true, citecolor=[rgb]{0, 0, 0.5}, linkcolor=[rgb]{0, 0, 0.5}, urlcolor=[rgb]{0, 0, 0.5}}
\usepackage{xstring}

\usepackage{color}

\usepackage{multirow}
\usepackage{booktabs}
\newcommand*\rot{\rotatebox{90}}
\newcommand{\linebreakcell}[2][c]{%
  \begin{tabular}[#1]{@{}c@{}}#2\end{tabular}}

\title{Evaluation of Transfer Learning for Polish with a Text-to-Text Model}

\name{Aleksandra Chrabrowa, Łukasz Dragan, Karol Grzegorczyk, Dariusz Kajtoch \\
\large{\textbf{Mikołaj Koszowski, Robert Mroczkowski, Piotr Rybak}}}

\address{Allegro ML Research at Allegro.pl, \\
         ul. Grunwaldzka 182, 60-166 Pozna\'{n}, Poland \\
         \{firstname.lastname\}@allegro.pl\\}

\abstract{
We introduce a new benchmark for assessing the quality of text-to-text models for Polish. The benchmark consists of diverse tasks and datasets: KLEJ benchmark adapted for text-to-text, en-pl translation, summarization, and question answering. In particular, since summarization and question answering lack benchmark datasets for the Polish language, we describe their construction and make them publicly available. Additionally, we present plT5 - a general-purpose text-to-text model for Polish that can be fine-tuned on various Natural Language Processing (NLP) tasks with a single training objective. Unsupervised denoising pre-training is performed efficiently by initializing the model weights with a multi-lingual T5 (mT5) counterpart. We evaluate the performance of plT5, mT5, Polish BART (plBART), and Polish GPT-2 (papuGaPT2). The plT5 scores top on all of these tasks except summarization, where plBART is best. In general (except for summarization), the larger the model, the better the results. The encoder-decoder architectures prove to be better than the decoder-only equivalent.
 \\ \newline  }

\begin{document}

\maketitleabstract

\section{Introduction}

Recent years have brought significant progress in natural language understanding (NLU) and natural language generation (NLG). Transformer architecture enabled efficient training of large-scale language models~\cite{radford2019language} and language understanding models~\cite{devlin-etal-2019-bert,liu2019roberta}. On the other hand, transfer learning, which has been used in representation learning for years~\cite{mikolov2013distributed,pennington-etal-2014-glove,devlin-etal-2019-bert}, has finally been successfully applied to text-to-text problems as well~\cite{raffel2020exploring,lewis-etal-2020-bart}. The text-to-text framework takes text as input and produces new text as output. This unified view enables the use of the same architecture, training procedure, and decoding process for many NLP tasks such as classification, machine translation, summarization, and question answering, to name a few. In addition, \newcite{raffel2020exploring} demonstrated that the simplicity of this approach combined with scale could achieve state-of-the-art results on many benchmark datasets, which makes it even more attractive. Their T5 model was available only for English language, but more recently pre-trained multi-lingual architectures~\cite{liu-etal-2020-multilingual-denoising,xue-etal-2021-mt5,nagoudi-etal-2021-indt5} and non-English  counterparts~\cite{carmo2020ptt5,Malaya} were released. Multiple publications show that models targeted for specific language perform better than multi-lingual one~\cite{martin-etal-2020-camembert,le-etal-2020-flaubert-unsupervised,chan-etal-2020-germans,mroczkowski-etal-2021-herbert,virtanen2019multilingual,nagoudi2021arat5}. Moreover, specialized architectures are typically smaller due to significantly reduced vocabulary size, and they can be trained efficiently via transfer from multi-lingual checkpoints~\cite{arkhipov-etal-2019-tuning,mroczkowski-etal-2021-herbert}. There were some attempts to pre-train Transformer-based models for generating Polish, namely plBART~\footnote{\url{https://github.com/sdadas/polish-nlp-resources\#bart}} and papuGaPT2~\footnote{\url{https://hf.co/flax-community/papuGaPT2}}, but they lack detailed description and evaluation on benchmark datasets. 

Our contributions are:
\begin{itemize}
    \item comprehensive evaluation of text-to-text models on diverse tasks in Polish, such as text-to-text KLEJ benchmark~\cite{rybak-etal-2020-klej}, machine translation, question answering and summarization,
    \item construction of benchmark datasets in the Polish domain for question answering and summarization,
    \item demonstration of the efficiency of pre-training procedure for transferring knowledge from multi-lingual to monolingual text-to-text models based on work by \cite{arkhipov-etal-2019-tuning,mroczkowski-etal-2021-herbert},
    \item release of plT5\footnote{\url{https://hf.co/allegro/plt5-base}} -- a T5-based model for the Polish language, which achieves the best results among the evaluated text-to-text models on KLEJ benchmark, machine translation and question answering and second-best results in summarization.
\end{itemize}

\section{Polish T5 model}

There was no large-scale text-to-text model for Polish to benchmark transfer learning methods, so we trained three versions of plT5 model (small, base, and large) with T5 architecture. We prepared a Polish language tokenizer, initialized the model weights based on mT5, and pre-trained it on Polish corpora.

\subsection{Architecture}
We followed the original mT5~\cite{xue-etal-2021-mt5} architectures with encoder and decoder stacks for all trained model sizes. We trained 3 variants of the T5 model: small (8 layers, 6 attention heads, hidden dimension 1024), base
(12 layers, 12 attention heads, hidden dimension 768) and large (24 layers, 16 attention heads, hidden dimension 1024).

\subsection{Initialization}
Parameters for all layers except word embeddings were copied directly from public mT5 models~\cite{xue-etal-2021-mt5}. We addressed the difference in tokenizers' vocabulary as described in~\cite{arkhipov-etal-2019-tuning,mroczkowski-etal-2021-herbert}. We copied embedding weights for tokens appearing in both dictionaries directly. Otherwise, if the target vocabulary token was missing in the source vocabulary, it was split into sub-tokens, and the embedding was the average of sub-tokens embeddings.

\subsection{Pre-training objective}
For the training of the plT5 model, we leveraged the self-supervised denoising objective proposed by~\cite{raffel2020exploring} i.e. span-corruption. The model was fed with the corrupted version of the original sentence during training. Single sentinel tokens replaced the consecutive spans of input tokens, and the target was to reconstruct replaced tokens.

\subsection{Pre-training datasets}

Denoising pre-training was performed on a weighted mixture\footnote{Probability of selecting given corpus in the mixture was proportional to the size of the dataset.} of the following corpora: Polish Wikipedia, National Corpus of Polish (NKJP, ~\newcite{przepiorkowski2011national}), Wolne Lektury\footnote{\url{https://wolnelektury.pl/}}, Polish Open Subtitles\footnote{\url{https://www.opensubtitles.org/pl}} and Common Crawl of Polish sites. Each dataset is described in~\cite{mroczkowski-etal-2021-herbert}. 

\subsection{Tokenizer}
The training dataset was tokenized into subword units using a sentencepiece unigram model~\cite{kudo-2018-subword} with a vocabulary size of 50k tokens. Unigram model was trained\footnote{\url{https://github.com/google/sentencepiece}} on the most representative parts of our corpus, i.e., the subset (random 25\% of the whole corpus) of the NKJP, and Polish Wikipedia.

\subsection{Pre-training procedure}
For self-supervised denoising pre-training, we used the original T5 training scripts\footnote{\url{https://github.com/google-research/text-to-text-transfer-transformer}},
which are based on the Mesh TensorFlow framework~\cite{shazeer2018mesh}. In addition, we used default options for the span-corruption objective with a mean span length of 3 and a corruption rate of 15\%~\cite{raffel2020exploring}. 

plT5 models were trained using AdaFactor optimizer~\cite{shazeer2018adafactor} with constant warmup and inverse square root learning rate schedule~\cite[section 3.1.2]{raffel2020exploring}. We use the peak learning rate 5e-3, batch of $1 048 576$ tokens, 0\% dropout, and trained for 50k steps with $1024$ warmup steps on a single preemptible TPU v3. 

The learning rate was found in a simple hypertuning procedure. We evaluated plT5-base model snapshots saved after 10k training steps for pre-training learning rates in \{1e-3, 2e-3, 5e-3\}. We used KLEJ benchmark tasks equipped with validation sets for performance evaluation: AR, NKJP-NER, PolEmo2.0-OUT, CDSC-E. Best model training was continued up to 50k steps.

\section{Downstream tasks}

plT5 is a generic text-to-text model for Polish. It can be fine-tuned on many NLP tasks. We evaluated it by fine-tuning on a text-to-text reformulation of the KLEJ benchmark, machine translation, question answering, and summarization. The same fine-tuning scheme was used for mT5 and other text-to-text or NLG models for Polish: plBART and papuGaPT2 to compare their performance. The trainable number of parameters and vocabulary size for each architecture is shown in Table~\ref{tab:model_stats}.

\begin{table}[ht]
    \centering
    \begin{tabular}{c|c|cc}
    \toprule
    \multicolumn{2}{c|}{\bf{Model}} & 
    \linebreakcell{\bf{\#}\\\bf{Params}} & \linebreakcell{\bf{Vocab.}\\\bf{size}}
    \\
    \toprule
    \textbf{Small} & mT5 & 300M & 250k \\
    \textbf{models} & plT5 & 95M & 50k
    \\
    \midrule
    & mT5 & 582M & 250k \\
    \textbf{Base} & plT5 & 275M & 50k \\
    \textbf{models} & plBART & 139M & 50k \\
    & papuGaPT2 & 124M & 50k
    \\
    \midrule
    \textbf{Large} & mT5 & 1.23B & 250k \\
    \textbf{models} & plT5 & 820M & 50k \\
    \bottomrule
    \end{tabular}
    \caption{Summary of the total number of trainable parameters for pre-trained models for Polish.}
    \label{tab:model_stats}
\end{table}

\subsection{General language understanding}

\label{par:klej_benchmark}
We verified the quality of the assessed models in terms of general language understanding with the KLEJ benchmark~\cite{rybak-etal-2020-klej}. We used all of the given tasks to compare the generative models, except for the CDSC-R regression. We formed the tasks into a text-to-text format proposed in~\cite{raffel2020exploring}. During the fine-tuning phase, the model was provided descriptive input containing the specific prefix and the textual features with descriptive labels\footnote{We did not explore multi-task learning in this setup with task-specific prompts}. The training objective was to generate greedily a specific label consisting of up to several tokens. Targets were generated over the whole vocabulary, and only an exact match was treated as the correct answer.
\subsubsection{Task definition}
We transformed KLEJ tasks into a text-to-text format where each input sentence (one or two in KLEJ tasks) has a prompt describing its role in the task semantics. Additionally, a task identification token could prefix whole input, but in all experiments in the paper, we did not use such a prompt because we were training models for one task at a time. Finally, labels were converted into semantically significant text tokens. Input features and task labels are in Table~\ref{tab:klej_t2t_formulation}.

\begin{table*}[ht]
\centering
\begin{tabular}{l|l|l|r}
\toprule
\textbf{Task} & \textbf{Prefix 1} & \textbf{Prefix 2} & \textbf{Labels}\\
\midrule
NKJP-NER  & zdanie  & - & geograficzna, brak, organizacja, osoba, miejsce, czas \\
CBD & zdanie & - & neutralna, przemoc  \\
Czy wiesz?  & pytanie  & odpowiedź & fałsz, prawda \\
PolEmo2.0 & zdanie & - & niejednoznaczny, negatywny, pozytywny, neutralny  \\
AR & zdanie & - & 1.0, 2.0, 3.0, 4.0, 5.0  \\
PSC & streszczenie 1 & streszczenie 2 & nie\_parafraza, parafraza  \\
CDSC-E & zdanie 1 & zdanie 2 & neutralny, wynikanie, sprzeczność  \\
\bottomrule
\end{tabular}
\caption{Prompts and labels used in the text-to-text formulation of KLEJ benchmark tasks. In case of encoder-decoder architectures, source was prepared as \texttt{<Prefix 1>}: text 1 \texttt{<Prefix 2>}: text 2. For decoder architecture, we additionally append \texttt{[SEP]} token at the end.}
\label{tab:klej_t2t_formulation}
\end{table*}

\subsubsection{Task evaluation}
At test time model is given input formatted as described above, and the goal is to generate a label. A sequence of target tokens is generated until EOS token generation or when it reaches max target length. We calculate the max target length per task, and it corresponds to the longest label after the tokenization process.
The generated sequence of tokens is converted into text. We count for a positive when it matches precisely the corresponding label.

\subsubsection{Experiments}
We followed a simple experimental setup to ensure a fair comparison between models. The learning rate was the only hyperparameter we tuned. We performed the best learning rate search for each model in the set of \{2e-5, 4e-5, 6e-5, 8e-5, 1e-4, 2e-4, 4e-4, 6e-4, 8e-4, 1e-3\}. For learning rate tuning, we used tasks with the validation set provided. The designated learning rate value was used for all test runs on all KLEJ tasks. By default, we trained the models for $1600$ steps with a batch size of $64$. Reported results are the median of 7 runs for small and base architectures and the median of 3 runs for the large ones.   

\subsubsection{Results}
The results for KLEJ benchmark (text-to-text format) are in Table \ref{tab:results-klej}. In general, encoder-decoder models dedicated to the Polish language perform significantly better at tasks in the KLEJ benchmark than their multi-lingual counterparts. The most striking performance difference occurs for the base version of the model on the PolEmo2.0-OUT task. The difference between plT5-base and mT5-base is over 9 pp. The difference on PolEmo2.0-OUT of more than 2pp is also visible for the other sizes. We hypothesize that domain adaptation necessary to solve this task reveals better Polish language understanding for monolingual T5. The superiority of the dedicated model is also visible in the PSC paraphrase identification task. Especially for the T5-small and T5-base models, the difference is 4.6pp and 3.9pp, respectively. The gap closes to 1pp for the large version of a T5 model. On the other hand, the multi-lingual mT5-large model obtained the best result on the NKJP-NER task, which shows that the task formulated in a classification form is more straightforward than its original version.

We observe a consistent increase in results with the size of the plT5 model in each task. The best encoder-decoder model is plT5-large. It performs the best in 6 out of the 8 evaluated tasks from the KLEJ benchmark. However, it is not as good as the best Polish encoder,  HerBERT-large. The difference between these architectures is, on average, 1pp. We observe the most significant performance gap of 6.8pp when moving from plT5-small to plT5-base. We found the biggest difference of around 20pp on \emph{PSC} and \emph{Czy wiesz?} tasks. 

Regarding the smaller architectures, the best text-to-text model is plT5-base. It is worth emphasizing that almost two times smaller plBART is very competitive. It is the best on 3 out of 8 tasks among the smaller models. The decoder only papuGaPT2 achieves on average results on par with plT5-small but is the worst at dealing with the highly unbalanced tasks, \emph{CBD} and \emph{Czy wiesz?}.

We emphasize the variability of the results for the two \emph{CBD} and \emph{Czy wiesz?} tasks. The difference between the results for different architectures is as high as 30pp.

To sum up, the models dedicated to the Polish language are the best from the perspective of the KLEJ benchmark. Among them, the T5 models were the best.

\begin{table*}[ht]
\renewcommand*{\arraystretch}{1.15}
\centering
\begin{tabular}{c|p{0.1\linewidth}|c|ccccccccc}

    \toprule
    \multicolumn{2}{c|}{\bf{Model}} & \rot{\bf{AVG }} & \rot{\bf{NKJP-NER }} & \rot{\bf{CBD}} & \rot{\bf{Czy wiesz? }} & \rot{\bf{PolEmo2.0-IN }} & \rot{\bf{PolEmo2.0-OUT }} & \rot{\bf{AR }} & \rot{\bf{PSC }} & \rot{\bf{CDSC-E }} 
    \\
    \toprule
    \textbf{Small} & \centering{mT5} & 74.3 $\pm$ 2.6 & 88.7 & 56.6 & 42.7 & 86.0 & 73.2 & 84.6 & 70.8 & 91.9  
    \\
    \textbf{models} & \centering{plT5} & \underline{76.8 $\pm$ 1.8} & \underline{92.4} & \underline{60.4} & \underline{44.7} & \underline{88.0} & \underline{76.6} & \underline{84.7} & \underline{75.4} & \underline{92.5}  
    \\
    \midrule
    & \centering{mT5} & 82.4 $\pm$ 0.0 & 92.8 & \underline{65.6} & 67.8 & 88.4 & 70.2 & 87.6 & 93.3 & 93.1 
    \\
    \textbf{Base} & \centering{papuGaPT2} & 76.5 $\pm$ 0.9 & 90.7 & 33.3 & 49.8 & 89.2 & 76.2 & 86.2 & 95.3 & 91.3  
    \\
    \textbf{models} & \centering{plBART} & 81.9 $\pm$ 0.5 & 93.1 & 47.6 & \underline{68.5} & 89.5 & 77.9 & \underline{88.0} &  \underline{97.9} & 93.2 
    \\
    & \centering{plT5} & \underline{83.4 $\pm$ 0.3} & \underline{93.6} & 62.3 & 63.8 & \underline{90.0} & \underline{79.3} & 87.8 & 97.2 & \underline{93.4}  
    \\
    & \centering{HerBERT} & 84.7 $\pm$ 0.4 & 94.5 & 66.4 & 64.3 & 90.9 & 80.4 & 87.7 & 98.9 & 94.5 
    \\
    \midrule
    \textbf{Large} & \centering{mT5} & 84.9 $\pm$ 6.7 & \textbf{\underline{94.8}} & 62.3 & \textbf{\underline{69.9}} & 91.4 & 80.3 & 88.8 & 97.9 & 93.5  
    \\
    \textbf{models} & \centering{plT5} & \bf{\underline{86.4 $\pm$ 0.3}} & 94.5 & \bf{\underline{70.0}} & 69.4 & \bf{\underline{92.9}} & \bf{\underline{82.6}} & \bf{\underline{88.9}} & \bf{\underline{98.9}} & \bf{\underline{94.0}} 
    \\
    & \centering{HerBERT} & 87.5 $\pm$ 0.2 & 96.4 & 72.0 & 75.8 & 92.2 & 81.8 & 89.1 & 98.9 & 94.1  
    \\
    \bottomrule

\end{tabular}
\caption{Evaluation results on KLEJ benchmark. \emph{AVG} is the average score across all tasks. Scores are reported for the test set and correspond to median values across seven runs for small and base models and three runs for large versions of models. The best scores for text-to-text models within each group are underlined. The best overall are in bold. Scores for the HerBERT model were taken from~\protect\cite{mroczkowski-etal-2021-herbert} and evaluated in a standard setting. F1 scores are reported for CBD, Czy wiesz? and PSC, Spearman correlations are reported for CDSC-R, MAE based score~\protect\cite{rybak-etal-2020-klej} for AR, and accuracy scores are reported for the other tasks.}
\label{tab:results-klej}
\end{table*}

\subsection{Summarization}
Summarization is the task of writing a shorter text that has all the key information of the longer text. Summaries can vary in length and be extractive (a selection of passages from the original text) or abstractive (written from scratch, though passages from the original text are not forbidden). 
\subsubsection{Datasets}
\paragraph{Allegro Articles (AA)} Collection of over 33k articles from Allegro.pl\footnote{Oryginal articles are available at \url{https://allegro.pl/artykuly}} - a popular e-commerce marketplace. They contain, among others, product reviews and shopping guides. Every article contains a title, lead (opening paragraph), and a body of text. We prepared 2 different summarization tasks\footnote{Prepared datasets are available at \url{https://hf.co/datasets/allegro/summarization-allegro-articles}}: (1) \emph{body2lead} - generate lead from a body of the article (2) \emph{body+lead2title} - generate a title from a full article (lead and body). The tasks are entirely abstractive and differ in source and target length. The details of dataset construction and statistics are in Appendix~\ref{subsec:aa_construction}.

\paragraph{Polish Summaries Corpus (PSC)} Collection of summaries for $569$ news articles created by human annotators~\cite{ogrodniczuk-kopec-2014-polish}. The annotators created 5 extractive summaries (each by different author) for each article by
choosing approximately $5\%$, $10\%$, or $20\%$ of the original text. The subset of $154$ articles was also supplemented with additional 5 abstractive summaries each, i.e., not created from the fragments of the original article. We constructed 3 summarization subsets\footnote{Prepared datasets are available at \url{https://hf.co/datasets/allegro/summarization-polish-summaries-corpus}}: (1) \emph{whole} - all summaries and articles (2) \emph{extract} - only extractive summaries and (3) \emph{abstract} - only abstractive summaries. The details of dataset construction and statistics are in Appendix~\ref{subsec:psc_construction}.

\subsubsection{Experiments}
To compare the models fairly, we designed an experimental setup where the learning rate is the only hyperparameter (details in Appendix~\ref{sub:lr_summary}). By default, each model was trained with three seeds for ten epochs with a batch size of 32 and Adam optimizer~\cite{kingma2014adam}. The only exceptions were large versions of T5 models (all tasks) and papuGaPT2 (PSC task) trained for four epochs.

\paragraph{Trimming inputs}\label{sub:trimming}
Summarization models by its nature deal with long source texts and shorter target texts. Unfortunately, we had to trim substantially source texts due to model and memory limitations. Usually, the limit was 1024 tokens, but sometimes it was even lower. The PSC tasks (news articles) were affected the most. The models never saw 100\% of source text during training, but between 26-52\%. Details are in Appendix~\ref{appendix:trimming}.
\subsubsection{Results}
Results are shown in Table~\ref{tab:summarization_results_all} which contains arithmetic mean of (f-measure) ROUGE-1, ROUGE-2 and ROUGE-L~\cite{lin-2004-ROUGE}\footnote{We used native Python implementation of ROUGE score~\url{https://github.com/google-research/google-research/tree/master/ROUGE} and replaced default tokenizer with nltk~\url{http://www.nltk.org/} punkt word tokenizer for Polish language.} for each model and task. More detailed metrics for each task can be found in Appendix~\ref{sec:summary_rouge} (Tables~\ref{tab:summarization_aa_body2lead}-\ref{tab:summarization_psc_abstract}). In all cases, the best performing model was plBART, with one exception of AA \emph{body+lead2title} task in which plT5-large outperformed the others. In terms of the average performance over all the tasks, plBART is the winner, and the second is plT5-small. plT5 models are always better than mT5 and papuGaPT2. We hypothesize that the pre-training procedure that makes BART biased towards copying the source is a good heuristic in the news summarization task. Furthermore, we observed that large architectures are unstable, and their performance degrades.
\paragraph{Baseline} To further evaluate the results, we computed 2 baselines: copying $n=3$ leading sentences from source as candidate target summarization, following~\cite{gliwa-etal-2019-samsum,see-etal-2017-get} and much better \emph{adaptive n} baseline: copying $n$ leading sentences from source that best match the average target length. The best model was usually 2-3pp above the adaptive baseline except for the PSC abstract task (0.1pp below the baseline) and AA \emph{body+lead2title} (32pp above the baseline).

\paragraph{Upper bound estimates} PSC dataset contains multiple summaries of different lengths and by different authors per source text. We evaluate human performance as ROUGE metrics between different summaries for the same source text. The best models are halfway between the baseline and upper bound for the PSC whole and PSC extract task. Interestingly, for PSC abstract, all: the baseline, the best model, and the upper bound are roughly the same. This shows that ROUGE is not a perfect metric, particularly for abstractive summaries. This is in agreement with the systematic evaluation of summarization metrics by \cite{summeval}.

\subsubsection{Discussion}

The PSC dataset contains summaries of different lengths that may degrade performance since the model may be uncertain about the length of the summary. One idea to lift that ambiguity is to add a prefix that would specify the target summary length. Furthermore, different source contexts seen by each model made interpretation of the results more challenging. If the source was trimmed down significantly, then the whole task turns into generation rather than summarization. On the other hand, copy baselines performed very well on PSC, and indeed, every model saw the leading sentences from the source during training.

After exploration, we found out that, on average, summaries generated by encoder-decoder architectures (plT5, mT5, plBART) are shorter than targets. The papuGaPT2 model generated summaries comparable in length to targets. We also spotted some anomalies. The mT5-large on PSC whole task generated summaries of maximal length, much longer than the target. Furthermore, the PSC abstract was particularly difficult for plT5-base which generated extremely short summaries (samples are in Appendix~\ref{sec:example_psc_whole},~\ref{sec:example_psc_extract},~\ref{sec:example_psc_abstract}).

\begin{table*}[ht]
    \centering
    \begin{tabular}{c|p{0.33\linewidth}|c|c|c|c|c|c} 
    \toprule
        \multicolumn{2}{c|}{\textbf{Model}} & 
        \linebreakcell{\textbf{ROUGE}\\\textbf{AVG}} & 
        \rot{\textbf{AA body2lead}} & 
        \rot{\textbf{AA body+lead2title}} & 
        \rot{\textbf{PSC whole}} & 
        \rot{\textbf{PSC extract}} & 
        \rot{\textbf{PSC abstract}} \\ 
        \toprule
        
        \multirow{2}{*}{\textbf{Baselines}} & \centering{lead n=3 source sentences} & 17.0 & \textbf{12.4} & 6.8 & 22.0 & 23.4 & 20.3 \\
        & \centering{lead n (adaptive) source sentences} & \textbf{21.6} & \textbf{12.4} & \textbf{7.9} & \textbf{30.3} & \textbf{31.7} & \textbf{25.6} \\

    \midrule
    \textbf{Upper bounds} & \centering{human performance} & - & - & - & \underline{34.3} & \underline{39.0} & 25.4 \\
        \midrule

    \multirow{2}{*}{\textbf{Small models}}& \centering{mT5} & \underline{23.3 $\pm$ 0.4} & \underline{13.0} & \underline{34.2} & 23.3 & 25.0 & 21.2 \\
    & \centering{plT5} & \underline{25.5 $\pm$ 6.2} & \underline{14.3} & \underline{36.3} & \underline{30.5} & 23.2 & 23.2 \\
    \midrule
        \multirow{4}{*}{\textbf{Base models}} & \centering{papuGaPT2} & 15.0 $\pm$  0.2 & 12.1 & 14.0 & 16.6 & 17.5 & 14.8 \\
         & \centering{mT5} & 20.2 $\pm$ 1.3 & 14.1 & 36.1 & 20.6 & 21.2 & 8.8 \\
        & \centering{plBART} & \textbf{\underline{29.3 $\pm$ 0.3}} & \textbf{\underline{15.6}} & \underline{38.3} & \textbf{\underline{32.6}} & \textbf{\underline{34.3}} & \textbf{25.5} \\
        & \centering{plT5} & \underline{23.1 $\pm$ 3.3} & \underline{14.9} & \underline{38.9} & 25.2 & 24.7 & 11.7 \\
        \midrule
         \multirow{2}{*}{\textbf{Large models}} & \centering{mT5} & 15.3 $\pm$ 1.1 & 10.9 & \underline{33.5} & 12.0 & 10.9 & 9.1\\
        & \centering{plT5} & 18.9 $\pm$ 1.8 & 12.1 & \textbf{39.4} & 17.5 & 15.1 & 10.6 \\
          \bottomrule
    \end{tabular}
    \caption{Evaluation results on all summarization tasks (test split, mean values across three runs). For simplicity only arithmetic mean ROUGE (f-measure) is reported, ROUGE-1, ROUGE-2 and ROUGE-L are in Appendix~\ref{sec:summary_rouge} (Tables~\ref{tab:summarization_aa_body2lead}-\ref{tab:summarization_psc_abstract}). The best scores are in bold, and results above the baseline are underlined. ROUGE AVG is the arithmetic mean of mean ROUGE of all tasks per model.}

    \label{tab:summarization_results_all}
\end{table*}

\subsection{Question answering}
Question answering (QA) systems enable users to automatically obtain accurate answers to natural language questions. We distinguish reading comprehension QA, in which the system, apart from the question, also receives a passage, which may contain the correct answer, and open-domain QA, in which the system receives only the question itself and answers only using the previously collected knowledge.

\subsubsection{Datasets}
There is no standard dataset for training the question answering system for the Polish language, so we combined several resources to create a QA task specifically for this evaluation.

\paragraph{MKQA} Multi-lingual Knowledge Questions \& Answers (MKQA)~\cite{mkqa} contains 10,000 queries sampled from the Google Natural Questions dataset~\cite{kwiatkowski-etal-2019-natural}, manually translated from English into 26 typologically diverse languages, including Polish.

After manually inspecting a sample of the data, we noticed many low-quality questions that cannot be answered using Wikipedia as a knowledge base. Therefore, we removed questions from specialized domains (e.g., TV series, movies)
with open-ended answers (''why?'' and ''how?'' questions), lacking the answer or the answer depending on when the question is asked (e.g. ''Who won World Cup this year?''). After filtration, we left 1875 valuable questions.

\paragraph{Jeden z Dziesięciu} is a Polish game show broadcast on Polish Television. The participants answer the host's questions from various domains. We gathered 1004 question-answer pairs\footnote{We parsed  \url{http://tvturnieje.blogspot.com/p/jeden-z-dziesieciu.html}}.

\paragraph{Poleval} is an evaluation campaign for NLP tools for Polish. The 2021 edition contains the question answering task\footnote{\url{https://github.com/poleval/2021-question-answering}} with a dataset of 6000 questions and answers. We used the validation set (1000 questions) for training and the test-A set (2500 questions) for the test. The test-B set was not released at the time of conducting the experiments.

\paragraph{Final dataset} combines the datasets mentioned above, which resulted in the training set of 3879 questions and the test set of 2500 questions.

\subsubsection{Tasks}
In the task of open-domain question answering, we aim to assess the factual knowledge stored in pre-trained models. Therefore, we expect that the evaluated models already have the required knowledge to answer the question, and we use fine-tuning procedure only to guide the models on how to generate the answer~\cite{10.1561/1500000001,roberts-etal-2020-much}.

The next question answering task is similar to the well-known reading comprehension task: besides the question, the model additionally takes a passage of text which may contain the answer~\cite{zeng2020survey}. Since none of the used datasets contains such passages, we retrieved them on our own in the following way.

First, we manually annotated 10k question-passage pairs with an information if the passage contains the correct answer. The source of candidates for positives pairs was severalfold. We started with a simple Bag-of-Words retriever as well as a more sophisticated model such as Universal Sentence Encoder~\cite{yang2019multilingual}. Next, we trained a neural retriever based on HerBERT-base model~\cite{mroczkowski-etal-2021-herbert} and annotated the retrieved passages.

Overall out of 10k annotated pairs, there were 2215 matching passages for 1347 unique questions\footnote{We release the dataset at \url{https://hf.co/datasets/allegro/polish-question-passage-pairs}
}.
We combined them with ''Czy wiesz?'' dataset~\cite{marcinczuk2013open,rybak-etal-2020-klej} to train a neural retriever based on the HerBERT-base model.

Finally, to prepare a dataset for the task, we used this retriever to find the top 10 best Wikipedia passages for all questions in our dataset. The goal of the task was to generate the answer based on a question and retrieved passages.

\subsubsection{Experiments} Using the aforementioned training set, we fine-tuned the following models: plT5-base, plT5-wiki\footnote{plT5-base additionally fine-tuned using Wikipedia corpus}, plT5-large, mT5-base, plBART, papuGaPT2, and T5 model initialized with random weights (T5-random) as a reference point. We trained the models for 50 epochs with Adam~\cite{kingma2014adam} optimizer except for plT5-large, which we trained for 30 epochs. We performed the best learning rate search for each model in the set of \{1e-2,5e-3, 1e-3, 5e-4\} with 1 seed value. We used a single NVIDIA A100 GPU for all the models. The best results for each model are presented in Table~\ref{tab:qa}. In both tasks, the best model was plT5-large. In base size, plT5-wiki performs the best in both tasks.

Models trained on open-ended tasks learned mainly to answer yes/no questions. In general, they were able to generate a reasonable and fluent response but mostly incorrect. Models also tend to overfit to the training examples from a similar domain (e.g., it answers ''Montmartre'' for the question of ''the highest peak of the Beskids range''). Fine-tuning the model on Wikipedia as the knowledge database improved the results on both tasks by 1.0pp and 0.3pp, respectively. However, directly providing retrieved passages improves the F1 score by 20pp. Thus, it is evident that there is still plenty of room to improve text-to-text models from a knowledge database perspective. 

In the passages subtask, we encountered a similar trimming issue as in the summarization task~\ref{sub:trimming}. Because of papuGaPT2 input size limitation, we could not pass the entire passage to the model. Therefore, we repeated plT5-wiki and plBART training using a 1024 input size setting. The results for plT5 and plBART were not significantly different from those with no size-limited input, which confirms their superiority over decoder-only architecture in this task.

\begin{table}[ht]
    \centering
    \begin{tabular}{c|c|c|c} 
        \toprule
        \multicolumn{2}{c|}{\textbf{Task}} & \textbf{open} & \textbf{passages} \\
        \toprule
        & plT5 & 20.9 & 42.8\\
        & plT5-wiki & \textbf{21.9} & \textbf{43.1}\\
        \textbf{Base models} & papuGaPT2 & 18.5 & 24.0\\
        & mT5 & 17.2 & 39.8\\
        & plBART & 17.4 & 37.1\\
        & T5-random & 10.9 & 8.0\\
        \midrule
         \textbf{Large models} & plT5 & \textbf{26.5} & \textbf{51.3}\\
          \bottomrule
    \end{tabular}
\caption{Accuracy scores for open questions and questions followed by the generated passages evaluated using Poleval \textit{testA} set.}
\label{tab:qa}
\end{table}

\subsection{Machine translation}

Machine translation is one of the most popular applications of text-to-text models. Therefore, we finetuned plT5 on parallel corpora consisting of pairs of English and Polish sentences and evaluated performance in both en$\rightarrow$pl and pl$\rightarrow$en directions.
\begin{table}[b]
\centering
\begin{tabular}{c|c|c|c}
\toprule
\multicolumn{2}{c|}{\textbf{Model}} & \textbf{Vocab} & \textbf{Batch size}\\
\toprule
                & mT5   & mT5                      & 25 \\
\textbf{Small}  & plT5  & plT5                     & 50 \\
\cmidrule{2-4}
\textbf{models} & mT5   & \multirow{2}{*}{wmt20}   & \multirow{2}{*}{50} \\
                & plT5  &                          & \\
\midrule
                & mT5   & mT5                      &  10 \\
                & plT5  & plT5                     &  25 \\
\cmidrule{2-4}
                & mT5   & \multirow{2}{*}{wmt20}   & \multirow{2}{*}{25} \\
\textbf{Base}   & plT5  &                          & \\
\cmidrule{2-4}
\textbf{models} & \multirow{2}{*}{plBART} & plBART & \multirow{2}{*}{40} \\
                &                         & wmt20  & \\
\cmidrule{2-4}
                & \multirow{2}{*}{\linebreakcell{papu-\\GaPT2}} & papuGaPT2 & \multirow{2}{*}{25} \\
                &                                               & wmt20     & \\
\midrule
\textbf{Large}  & mT5  & \multirow{2}{*}{wmt20} & \multirow{2}{*}{10} \\
\textbf{models} & plT5 &                        & \\
\bottomrule
\end{tabular}
\caption{Machine translation batch sizes}
\label{tab:wmt_batch_sizes}
\end{table}

\subsubsection{Datasets}

We used datasets collected for the news translation competition organized as part of the 5th Conference on Machine Translation\footnote{\url{https://www.statmt.org/wmt20/translation-task.html}} (WMT20). Specifically, we used 4 out of 5 parallel corpora available to the competition’s participants: EuroParl v10, TildeRapid, WikiTitles v2, and ParaCrawl v5.1. The corpora are described by~\newcite{barrault-etal-2020-findings}. We did not use the 5th dataset,  WikiMatrix, which is known to introduce more noise than useful knowledge~\cite{caswell2021quality}. To all 4 corpora, we applied similar filtering as~\newcite{DBLP:conf/tsd/JonssonSSSL20}.

\subsubsection{Experiments}

We evaluated the models on the WMT20 development set using the BLEU score~\cite{papineni-etal-2002-bleu}. For the generation, we used beam search with five beams and a maximal sequence length of 100. Input sentences were also limited to 100 tokens. We compared plT5, mT5, plBART, and papuGaPT2. The models were trained in both directions simultaneously, with examples fed to the models alternately. Since the plT5 tokenizer was trained on Polish-only data, we trained a separate tokenizer for WMT20 data. It was a unigram language model~\cite{kudo-2018-subword} with a vocabulary of 32k tokens based on our training corpora with the same filtering. In order to initialize embeddings of tokens from the WMT20 tokenizer that are not present in the plT5 tokenizer, we applied a technique based on~\newcite[section 3]{arkhipov-etal-2019-tuning}. We applied the same conversion to those models to have a fair comparison with mT5, plBART, and papuGaPT2. To distinguish between en$\rightarrow$pl and pl$\rightarrow$en directions, we prepended source sentences with either \texttt{<2en>} or \texttt{<2pl>} special tokens. 

\begin{table}[b]
\centering
\begin{tabular}{c|c|c|ll}
\toprule
\multicolumn{2}{c|}{\textbf{Model}} & \textbf{Vocab} & \rot{\textbf{en$\rightarrow$pl\ }} & \rot{\textbf{pl$\rightarrow$en\ }}\\
\toprule
                 & mT5  & mT5                       & 20.5             & 25.0             \\
\textbf{Small}   & plT5 & plT5                      & 20.3             & 24.7             \\
\cmidrule{2-5}
\textbf{models}  & mT5  & \multirow{2}{*}{wmt20}    & \underline{20.8} & 24.8             \\
                 & plT5 &                           & \underline{20.8} & \underline{25.4} \\
\midrule
                 & mT5  & mT5                       & 22.4             & \underline{27.0} \\
                 & plT5 & plT5                      & \underline{23.2} & 26.7             \\
\cmidrule{2-5}
                 & mT5  & \multirow{2}{*}{wmt20}    & 22.5             & 26.7             \\
\textbf{Base}    & plT5 &                           & 22.7.            & 26.8             \\
\cmidrule{2-5}
\textbf{models}  & \multirow{2}{*}{plBART} & plBART & 21.2             & 25.4             \\
                 &                         & wmt20  & 21.6             & 26.3             \\
\cmidrule{2-5}
                 & \multirow{2}{*}{\linebreakcell{papu-\\GaPT2}} & papuGaPT2 & 21.2 & 25.5 \\
                 &                                               & wmt20     & 22.0 & 26.2 \\
\midrule
\textbf{Large}   & mT5  & \multirow{2}{*}{wmt20} & 24.8                      & \textbf{\underline{29.0}} \\
\textbf{models}  & plT5 &                        & \textbf{\underline{25.5}} & 28.9                      \\
\bottomrule
\end{tabular}
\caption{BLEU scores for the WMT20 devset. The best results in each category (small, base, and large) are underlined. The best overall results are shown in bold.}
\label{tab:wmt20}
\end{table}

For training, we used Adam optimizer~\cite{kingma2014adam} with learning rate 1e-1 for small models and 1e-2 for the base and large ones, gradient accumulation of 8 batches, 10k steps of a linear warm-up, and inverse square root learning rate schedule~\cite[section 3.1.2]{raffel2020exploring}. We used different batch sizes for different model sizes to accommodate as many sentences in a batch as possible. Specific batch sizes are listed in Table~\ref{tab:wmt_batch_sizes}. We used a single NVIDIA A100 GPU with 40 GB of memory for all the models. Small and base models were trained for five epochs and large for two.

\subsubsection{Results}

The results are presented in Table~\ref{tab:wmt20}. plT5 generates translations superior to mT5 in direction en$\rightarrow$pl but not in pl$\rightarrow$en. This may be because plT5 was trained on Polish-only data and, therefore, the ability to generate English text deteriorates. plT5 also outperforms plBART and papuGaPT2. We repeated plT5-base and plBART experiments for three different seeds. The variance of the plT5 results was up to 0.1 BLEU, while the variance of the plBART results was slightly higher, up to 0.5 BLEU.

It should be stated that the results reported for plT5 are not up to the level of the WMT20 winning team's results~\cite{krubinski-etal-2020-samsung}. However, plT5-large performance in the en$\rightarrow$pl direction is slightly better than their bi-directional baseline. However, in this work, we did not attempt to achieve state-of-the-art on those datasets but to compare plT5 with similar architectures in similar settings.

\section{Conclusion}

In this work, we introduced a benchmark for natural language generation in Polish composed of 4 tasks constructed from publicly available datasets, namely the KLEJ benchmark adapted for text-to-text, en-pl translation, summarization with varying levels of abstractedness, and question answering. In addition, we introduced a novel text-to-text model for Polish, plT5. It outperforms mT5 on the KLEJ benchmark, summarization, en-pl machine translation, and question answering. It is better than plBART and papuGaPT2, except for summarization, where plBART is the best. We attribute this performance to the pre-training scheme where BART reconstructs the entire input in the decoder and thus possesses \emph{copy bias}. We should state that the overall performance of plBART is impressive, considering that it has almost twice as few parameters as plT5-base.
We observed that the larger the model, the better the results (except the summarization), and encoder-decoder architectures are better than decoder only. We efficiently pre-trained plT5 by initializing the weights from mT5 checkpoints with no exhaustive training. We made our models and new generation benchmark publicly available.

We did not explore the possibility of optimizing prefixes and labels for the text-to-text KLEJ benchmark and leave for future work. Thus, results reported in Table.~\ref{tab:results-klej} need to be treated as the lower bound. Additionally, our summarization dataset includes summaries of different lengths. This introduces additional ambiguity, especially during evaluation, since the model does not know what should be the target length of the summary. It would be interesting to check if conditioning the text-to-text model on summary length would lead to better results.

\section*{Acknowledgments}
We thank Alina Wróblewska from Institute of Computer Science, Polish Academy of Sciences for help with preprocessing the National Corpus of Polish.

\section*{References}

\bibliographystyle{lrec2022-bib}
\bibliography{acl,custom}

\clearpage
\appendix


\section{Summary of the released datasets}\label{sec:datasets_summary}

We released six new datasets. All of them are available at \url{https://hf.co/allegro}. Their sizes are shown in Table~\ref{tab:datasets_summary}.

\begin{table}[hbt!]
    \centering
    \begin{tabular}{c|c|c|c} 
    \toprule
    Dataset            & \multicolumn{3}{c}{Examples} \\
    name               & Train  & Dev   & Test        \\ \midrule
    AA body2lead       & 380351 & 41966 & 104514      \\
    AA body+lead2title & 380351 & 41966 & 104514      \\
    PSC whole          & 512201 & 57180 & 141293      \\
    PSC extract        & 413139 & 45413 & 112880      \\
    PSC abstract       & 100043 & 11222 & 27976       \\
    Polish Q-P Pairs.  & 10429  &       &             \\
    \bottomrule
    \end{tabular}
    \caption{Sizes of the released datasets.}
    \label{tab:datasets_summary}
\end{table}

\section{Summarization task}
   \paragraph{Level of abstractedness}
   Abstractedness of the dataset is measured by calculating the unique n-grams in the reference summary, which are not in the article~\cite{koupaee2018wikihow}.
   \paragraph{Compression ratio}
   The compression ratio is defined as the ratio between the average length of the source and the average length of summaries.
  \subsection{Allegro Reviews (AA)}\label{subsec:aa_construction}
    The raw dataset is a collection of articles scrapped from \url{https://allegro.pl/artykuly} website. Each example contains title, lead, body, information about the category tree, and other metadata. In the pre-processing stage, we removed markdown formatting and normalized white spaces. This dataset serves as a good benchmark for highly abstractive summarization/generation tasks since every article contains a title or lead that can be viewed as a loose abstract of the article. Moreover, every article is written by a professional editor, and they are relatively short hence fit almost entirely into 512 token context. 
    
     We constructed two versions of the abstractive summarization task, which affect the summary ratio. In \emph{body2lead} task, we use the body of the article as a source and lead as a generation target. In \emph{body+lead2title} we use concatenated lead and body as source (full article) and generate title as a target. 
  
     For evaluation purposes, we divided the whole data source into a $80\%$ train set and $20\%$ test set in a stratified way using a top-level category in the Allegro category tree as a label. Then, we saved $10\%$ of the train set for validation (also stratified split) and hyperparameter tuning. Table~\ref{tab:summarization_datasets} contains summary of average source/target length and size of each split.

  \subsection{Polish Summaries Corpus (PSC)}\label{subsec:psc_construction}
  Original dataset~\cite{ogrodniczuk-kopec-2014-polish} contains extractive or abstractive summaries and other metadata for news articles prepared by annotators\footnote{We used publicly available version \url{https://hf.co/datasets/polsum}}. Each article has a few corresponding summaries that depend on the annotator, summary ratio (5, 10, or 20\% of the original text), and type (extractive or abstractive). Due to the small size of the entire dataset, we included all summaries in the final dataset. In this way, the source contains duplicates and targets near-duplicates. Moreover, our train set contains target summaries of different ratios that may affect learning and prediction. 
  We constructed 3 versions of the summarization task. The \emph{extract} subset contains only extractive summaries, the \emph{abstract} subset contains only abstractive summaries and \emph{whole} contains both of them.
  
  For evaluation purposes, we divided the whole data source into $80\%$ train set and $20\%$ test set in a stratified way using summary ratio, summary type, and article category as a label. Then, we saved $10\%$ of the train set for validation (also stratified split) and hyperparameter tuning. Table~\ref{tab:summarization_datasets} contains summary of average source/target length and size of each split.

\begin{table}[ht]
\centering
\begin{tabular}{c|cc|ccc} \toprule
\rot{\bf{Task}} & \rot{\bf{AA body2lead}} & \rot{\bf{AA body+lead2title}} & \rot{\bf{PSC whole}} & \rot{\bf{PSC extract}} & \rot{\bf{PSC abstract}} \\
\midrule
        \textbf{Domain} & \multicolumn{2}{c|}{E-commerce} & \multicolumn{3}{c}{News articles} \\
        \midrule
        \multicolumn{6}{c}{\textbf{Average length (characters)}} \\
        \midrule
        \textbf{Source} & 3.9k & 4.1k & \multicolumn{3}{c}{10.6k} \\ 
        \textbf{Target} & 0.3k & 0.05k & \multicolumn{3}{c}{1.3k} \\
        \midrule
        \multicolumn{6}{c}{\textbf{Compression ratio}} \\
        \midrule
        \textbf{-} & 13.0 & 82.0 & \multicolumn{3}{c}{8.2} \\
        \midrule
        \multicolumn{6}{c}{\textbf{Level of abstractedness}} \\
        \midrule
        \textbf{1-grams} & 0.53 & 0.30 & 0.05 & 0.01 & 0.23 \\
        \textbf{2-grams} & 0.87 & 0.69 & 0.16 & 0.07 & 0.48 \\
        \textbf{3-grams} & 0.95 & 0.85 & 0.25 & 0.15 & 0.61 \\
        \textbf{4-grams} & 0.97 & 0.92 & 0.32 & 0.22 & 0.68 \\
        \textbf{5-grams} & 0.98 & 0.95 & 0.38 & 0.28 & 0.73 \\
        \midrule
        \multicolumn{6}{c}{\textbf{Number of examples}} \\
        \midrule
        \textbf{Train} & \multicolumn{2}{c|}{24.4k} & 7.8k & 6.1k & 1.7k \\ 
        \textbf{Dev} & \multicolumn{2}{c|}{2.7k} & 0.9k & 0.7k & 0.2k \\ 
        \textbf{Test} & \multicolumn{2}{c|}{6.8k} & 2.2k & 1.7k & 0.5k \\ 
        \bottomrule

\end{tabular}
\caption{Overview of summarization tasks with the total number of examples for each split and average length (in characters) of source and target}

\label{tab:summarization_datasets}
\end{table}
\subsection{Trimming inputs}\label{appendix:trimming} The BART and GPT-2 architectures use fixed positional encoding, which is limited to 1024 tokens by default. On the other hand, the T5 model uses relative positional encoding and can process sequences of any length in principle. In our case, we are bounded by memory requirements. Thus, we fixed 1024 tokens as the maximal sequence length for all models. Even then, large architectures required further trimming of input tensors due to memory errors. Consequently, we could not fit the whole target and source texts into the model inputs.

Since each model uses a different tokenizer, the context used during training and prediction was different for individual models. The effect is more pronounced for source texts because they are much longer than the targets. We were always able to input 98-100\% of target tokens into the model and assumed that target trimming did not substantially affect the results. Usually, only a few outliers (extremely long targets) were trimmed. 

On the other hand, source trimming could have affected the models. The final number of tokens to which the source was trimmed and the percentage of the source tokens seen by the model are shown in Table~\ref{tab:summarization_trim}. The PSC task was affected the most. For example, mT5-large saw, on average only 25\% of the source. Similarly, the papuGaPT2 saw, on average only 26\% of the source. Less affected were plBART and plT5 models, which saw above 50\% of the source. Most importantly, none of the models ever saw 100\% of source texts during training for PSC tasks.

\begin{table}[ht]
    \centering
    \begin{tabular}{c|c|c|c|c} 
        \toprule
        \multicolumn{2}{c|}{\textbf{Task}} & \rot{\textbf{PSC}} & \rot{\textbf{AA body2lead}} & \rot{\textbf{AA body+lead2title}} \\
        \toprule
         \textbf{Small} & mT5 & 33\% & 85\% &  80\% \\
         \textbf{models} & plT5 & 51\% & 96\% &  95\% \\
    \midrule
         & papuGaPT2 & \linebreakcell{26\%\\(512)} & \linebreakcell{92\%\\(767)} & \linebreakcell{96\%\\(959)} \\
         \textbf{Base} & mT5 & 33\% & 85\% &  80\% \\
        \textbf{models} & plBART & 52\% & 96\% &  95\% \\
         & plT5 & 51\% & 96\% & 95\% \\
        \midrule
         \linebreakcell{\textbf{Large}\\\textbf{models}} & mT5 & \linebreakcell{25\%\\(768)} & \linebreakcell{85\%\\\ } &  \linebreakcell{80\%\\\ }\\
         & plT5 & 51\% & 96\% & 95\% \\
          \bottomrule
    \end{tabular}
    \caption{Average percentage of tokens from source text that fit into model. Differences are due to a tokenizer, model capacity, and memory limitations. By default, source text was trimmed to 1024 tokens, and the exceptions are indicated (in parentheses) in the table.}
    \label{tab:summarization_trim}
\end{table} 

  \subsection{Learning rate tuning}\label{sub:lr_summary}
  
  On the AA task, we performed best learning rate search in the range \{1e-5, 2e-5, 4e-5, 6e-5, 8e-5, 1e-4, 2e-4, 4e-4, 6e-4, 8e-4, 1e-3\}. On the PSC task, we performed a learning rate search for each model on the PSC whole subset and used the same learning rates for the PSC extract and PSC abstract subsets. The search was geometrical, with the start point at 4e-4, i.e., at the first subset \{2e-4, 4e-4, 8e-4\} was examined, each learning rate with three seed runs. The search was completed if the middle point was better than the two others. Otherwise either \{1e-4, 2-e4, 4e-4\} or \{4e-4, 8e-4, 1.6e-3\} learning rates were examined during the next step depending on the trend. As a result, learning rates from 5e-5 to 2.56e-2 were selected for different models. The only exception was the mT5-large model, which was unstable and required finer granularity \{2.5e-5, 5e-5, 7.5e-5 \} was used at the end of the search. Our procedure of learning rate tuning was motivated by budget constraints.

\section{Summarization ROUGE scores}\label{sec:summary_rouge}
  \begin{table}[ht]
    \centering
    \begin{tabular}{c|p{0.2\linewidth}|c|c|c|c}
    \toprule
        \multicolumn{2}{c|}{\textbf{Model}} & \rot{\textbf{ROUGE-1}} & \rot{\textbf{ROUGE-2}} & \rot{\textbf{ROUGE-L}} & \rot{\textbf{ROUGE}} \\ 
        \toprule
        
        \textbf{Baselines} & \centering{n=3} & \textbf{20.3} & \textbf{2.8} & \textbf{14.1} & \textbf{12.4} \\
        \linebreakcell{lead source\\sentences} & \centering{adaptive}& 20.2 & \textbf{2.8} & \textbf{14.1} & \textbf{12.4} \\
    \midrule
         \textbf{Small} & \centering{mT5} & 19.7 & \underline{4.1} & \underline{15.2} & \underline{13.0} \\
         \textbf{models} & \centering{plT5} & \underline{21.5} & \underline{4.9} & \underline{16.5} & \underline{14.3} \\
    \midrule
         & \centering{papuGaPT2} & 20.2 & 1.6 & 12.6 & 12.1 \\
        \textbf{Base} & \centering{mT5} & \underline{21.4} & 4.6 & \underline{16.3} & \underline{14.1} \\
         \textbf{models} & \centering{plT5} & \underline{22.3} & \underline{5.2} & \underline{17.1} & \underline{14.9} \\
         & \centering{plBART} & \underline{\textbf{24.0}} & \underline{\textbf{5.1}} & \underline{\textbf{17.7}} & \underline{\textbf{15.6}} \\

         \midrule
         \textbf{Large} & \centering{mT5} & 16.7 & 2.7 & 13.4 & 10.9\\
         \textbf{models} & \centering{plT5} & 19.2 & 2.7 & \underline{14.4} & 12.1 \\
          \bottomrule
    \end{tabular}
    \caption{Evaluation results on Allegro Articles body2lead task (test split, mean values across 3 runs). The best scores are in bold, results above the baseline are underlined. The ROUGE score reported in the last column is arithmetic mean of ROUGE-1, ROUGE-2 and ROUGE-L scores. All reported ROUGE scores are f-measures.}
    \label{tab:summarization_aa_body2lead}
\end{table}

\begin{table}[ht]
    \centering
    \begin{tabular}{c|p{0.2\linewidth}|c|c|c|c}
    \toprule
        \multicolumn{2}{c|}{\textbf{Model}} & \rot{\textbf{ROUGE-1}} & \rot{\textbf{ROUGE-2}} & \rot{\textbf{ROUGE-L}} & \rot{\textbf{ROUGE}} \\ 
        \toprule
        
        \textbf{Baselines} & \centering{n=3} & 9.3 & 2.9 & 8.1 & 6.8 \\
        \linebreakcell{lead source\\sentences} & \centering{adaptive} & \textbf{10.4} & \textbf{3.6} & \textbf{9.7} & \textbf{7.9} \\
        \midrule
        
        \textbf{Small} & \centering{mT5} & \underline{39.3} & \underline{25.1} & \underline{38.2} & \underline{34.2} \\
        \textbf{models} & \centering{plT5} & \underline{41.4} & \underline{27.2} & \underline{40.1} & \underline{36.3} \\
             \midrule
        & \centering{papuGaPT2} & \underline{18.8} & \underline{5.2} & \underline{18.0} & \underline{14.0} \\
        & \centering{mT5} & \underline{41.3} & \underline{27.0} & \underline{40.1} & \underline{36.1} \\
        \textbf{Base}  & \centering{plBART} & \underline{44.0} & \underline{29.6} & \underline{41.3} & \underline{38.3} \\
        \textbf{models} & \centering{plBART$^\dagger$} & \underline{44.0} & \underline{29.7} & \underline{42.4} & \underline{38.7} \\
         & \centering{plT5} & \underline{44.1} & \underline{29.7} & \underline{42.8} &  \underline{38.9} \\
         & \centering{plT5$^\dagger$} & \underline{43.9} & \underline{29.4} & \underline{42.6} &  \underline{38.7} \\
         \midrule
         \textbf{Large} & \centering{mT5} & \underline{38.4} & \underline{24.7} & \underline{37.3} & \underline{33.5}\\
         \textbf{models} & \centering{plT5} & \underline{\textbf{44.6}} & \underline{\textbf{30.2}} & \underline{\textbf{43.3}} & \underline{\textbf{39.4}} \\
          \bottomrule
    \end{tabular}
    \caption{Evaluation results on Allegro Articles body+lead2title task (test split, mean values across 3 runs). The best scores are in bold, results above the baseline are underlined. The ROUGE score reported in the last column is arithmetic mean of ROUGE-1, ROUGE-2 and ROUGE-L scores. All reported ROUGE scores are f-measures. ($\dagger$) models were trained with the same input and target truncation as papuGaPT2.}
    \label{tab:summarization_aa_body+lead2title}
\end{table}

\begin{table}[ht]
    \centering
    \begin{tabular}{c|p{0.2\linewidth}|c|c|c|c}
    \toprule
        \multicolumn{2}{c|}{\textbf{Model}} & \rot{\textbf{ROUGE-1}} & \rot{\textbf{ROUGE-2}} & \rot{\textbf{ROUGE-L}} & \rot{\textbf{ROUGE}} \\ 
        \toprule
        \textbf{Baselines} & \centering{n=3} & 28.2 & 16.5 & 21.3 & 22.0 \\
        \linebreakcell{lead source\\sentences} & \centering{adaptive} & \textbf{39.0} & \textbf{23.8} & \textbf{28.3} & \textbf{30.3} \\
    \midrule
    \textbf{\linebreakcell{Upper\\bounds}} & \centering{\linebreakcell{human\\perfor-\\mance}} & \underline{41.8} & \underline{27.4} & \underline{33.8} & \underline{34.3} \\
    \midrule
        \textbf{Small} &  \centering{mT5} & 29.4 & 17.0 & 23.5 & 23.3\\
         \textbf{models} & \centering{plT5} & 36.7 & \underline{24.3} & \underline{30.4} & \underline{30.5} \\
             \midrule
         &  \centering{papuGaPT2} & 26.9 & 7.6 & 15.5 & 16.7 \\
         \textbf{Base} &\centering{mT5} & 27.4 & 14.1 & 20.3 & 20.6 \\
         \textbf{models} & \centering{plBART} & \textbf{39.0} & \underline{\textbf{26.3}} & \underline{\textbf{32.4}} & \underline{\textbf{32.6}} \\
         & \centering{plT5} & 32.9 & 18.1 & 24.5 &  25.2 \\
         \midrule
         \textbf{Large} &  \centering{mT5} & 20.6 & 3.7 & 11.5 & 12.0\\
         \textbf{models} &  \centering{plT5} & 23.1 & 12.0 & 17.4 & 17.5\\
          \bottomrule
    \end{tabular}
    \caption{Evaluation results on Polish Summary Corpus (whole) task (test split, mean values across 3 runs). The best scores are in bold, results above the baseline are underlined. The ROUGE score reported in the last column is arithmetic mean of ROUGE-1, ROUGE-2 and ROUGE-L scores. All reported ROUGE scores are f-measures.}
    \label{tab:summarization_psc_full}
\end{table}

\begin{table}[ht]
    \centering
    \begin{tabular}{c|p{0.2\linewidth}|c|c|c|c}
    \toprule
        \multicolumn{2}{c|}{\textbf{Model}} & \rot{\textbf{ROUGE-1}} & \rot{\textbf{ROUGE-2}} & \rot{\textbf{ROUGE-L}} & \rot{\textbf{ROUGE}} \\
        \toprule

        \textbf{Baselines} & \centering{n=3} & 29.5 & 18.3 & 22.6 & 23.4 \\
        \linebreakcell{lead source\\sentences} & \centering{adaptive}& \textbf{40.0} & \textbf{25.5} & \textbf{29.7} & \textbf{31.7} \\
    \midrule
        \textbf{\linebreakcell{Upper\\bounds}} & \centering{\linebreakcell{human\\perfor-\\mance}} & \underline{45.1} & \underline{33.3} & \underline{38.8} & \underline{39.0} \\
        \midrule
         \textbf{Small} & \centering{mT5} & 30.4 & 19.3 & 25.3 & 25.0\\
         \textbf{models} & \centering{plT5} & 27.7 & 18.3 & 23.6 & 23.2\\
             \midrule
         & \centering{papuGaPT2} & 26.9 & 8.5 & 15.9 & 17.1\\
         \textbf{Base} & \centering{mT5} & 27.6 & 15.0 & 20.9 & 21.2\\
         \textbf{models} & \centering{plBART} & \textbf{39.9} & \underline{\textbf{28.6}} & \underline{\textbf{34.4}} & \underline{\textbf{34.3}} \\
         & \centering{plT5} & 31.9 & 18.0 & 24.1 &  24.7\\
         \midrule

         \textbf{Large} & \centering{mT5} & 19.3 & 2.6 & 10.9 & 10.9\\
         \textbf{models} & \centering{plT5} & 22.4 & 7.3 & 15.6 & 15.1\\
          \bottomrule
    \end{tabular}
    \caption{Evaluation results on Polish Summary Corpus (extract) task (test split, mean values across 3 runs). The best scores are in bold, results above the baseline are underlined. The ROUGE score reported in the last column is arithmetic mean of ROUGE-1, ROUGE-2 and ROUGE-L scores. All reported ROUGE scores are f-measures.}
    \label{tab:summarization_psc_extract}
\end{table}

\begin{table}[ht]
    \centering
    \begin{tabular}{c|p{0.2\linewidth}|c|c|c|c}
    \toprule
        \multicolumn{2}{c|}{\textbf{Model}} & \rot{\textbf{ROUGE-1}} & \rot{\textbf{ROUGE-2}} & \rot{\textbf{ROUGE-L}} & \rot{\textbf{ROUGE}} \\ 
        \toprule
        \textbf{Baselines} & \centering{n=3} & 27.1 & 14.2 & 19.7 & 20.3 \\
        \linebreakcell{lead source\\sentences} & \centering{adaptive} & \textbf{35.4} & \textbf{17.8} & \textbf{23.6} & \textbf{25.6} \\
    \midrule
        \textbf{\linebreakcell{Upper\\bounds}} & \centering{\linebreakcell{human\\perfor-\\mance}} & 34.6 & 16.7 & \underline{25.0} & 25.4 \\
        \midrule
         \textbf{Small} & \centering{mT5} & 28.0 & 14.9 & 20.7 & 21.2\\
         \textbf{models} & \centering{plT5} & 30.4 & 16.5 & 22.7 & 23.0\\
             \midrule
         & \centering{papuGaPT2} & 26.0 & 5.7 & 13.9 & 15.2\\
         \textbf{Base} & \centering{mT5} & 12.0 & 5.0 & 9.5 & 8.8\\
         \textbf{models} & \centering{plBART} & \underline{\textbf{33.5}} & \textbf{17.7} & \underline{\textbf{25.4}} & \textbf{25.5} \\
         & \centering{plT5} & 15.6 & 7.7 & 12.0 &  11.7\\
         \midrule
         \textbf{Large} & \centering{mT5} & 16.1 & 1.9 & 9.4 & 9.1\\
         \textbf{models} & \centering{plT5} & 15.6 & 5.1 & 11.2 & 10.6\\
          \bottomrule
    \end{tabular}
    \caption{Evaluation results on Polish Summary Corpus (abstract) task (test split, mean values across 3 runs). The best scores are in bold, results above the baseline are underlined. The ROUGE score reported in the last column is arithmetic mean of ROUGE-1, ROUGE-2 and ROUGE-L scores. All reported ROUGE scores are f-measures. }
    \label{tab:summarization_psc_abstract}
\end{table}

\onecolumn
\clearpage
\section{Example predictions on Allegro Reviews \emph{body+lead2title}}
\begin{enumerate}
    \item \textbf{Gold Summary:} Speedminton – badminton na wietrzne dni 
    
    \textbf{plT5-base:} Jak zacząć grać w speedminton?

    \textbf{plBART:} Speedminton – sport dla każdego
 
    \textbf{papuGaPT2:} Squa – jak zacząć?
    
    \textbf{mT5-base:} Jak zacząć grać w speedmintona?
    
    \textbf{plT5-large:} Speedminton – sport dla aktywnych
    
    \item \textbf{Gold Summary:} Urządzamy wnętrze po skandynawsku

    \textbf{plT5-base:} Jak urządzić mieszkanie w stylu skandynawskim?

    \textbf{plBART:} Jak urządzić mieszkanie w stylu skandynawskim?

    \textbf{papuGaPT2:} Skandynawski styl w wersji budżetowej – jak go urządzić?

    \textbf{mT5-base:} Jak urządzić wnętrze w stylu skandynawskim? 
    
    \textbf{plT5-large:} Jak urządzić mieszkanie w stylu skandynawskim?
    
    \item \textbf{Gold Summary:} Czy warto robić domowe peelingi – poradnik

    \textbf{plT5-base:} Najlepsze peelingi do domu

    \textbf{plBART:} Najlepsze peelingi do twarzy

    \textbf{papuGaPT2:} Peeslingi – najlepszy sposób na przesuszoną skórę

    \textbf{mT5-base:} Jak zrobić peelingi w domu?
 
    \textbf{plT5-large:} Jak zrobić peeling w domu?
        
    \item \textbf{Gold Summary:} Szerokie spodnie – must have sezonu wiosna/lato 2015

    \textbf{plT5-base:} Wiosenne trendy – szerokie spodnie

    \textbf{plBART:} Modne szerokie spodnie – jak je nosić?

    \textbf{papuGaPT2:} Wiosenne stylizacje ze schullami w roli głównej

    \textbf{mT5-base:} Wiosenny trend – szerokie spodnie
 
    \textbf{plT5-large:} Wiosenne spodnie w stylu lat 70.

    \item \textbf{Gold Summary:} Przygotuj się na lodowisko

    \textbf{plT5-base:} Jak rozpocząć sezon na łyżwy?

    \textbf{plBART:} Jak zacząć przygodę z łyżwami?

    \textbf{papuGaPT2:} Jakie ubrania na sezon kąpielowy?

    \textbf{mT5-base"} Jak zacząć sezon na łyżwy?
 
    \textbf{plT5-large:} Jak przygotować się do sezonu na łyżwy?

    \item \textbf{Gold Summary:} Wózek widłowy niczym limuzyna sterowana iPadem

    \textbf{plT5-base:} Wózki widłowe zwiększające możliwości swojego działania

    \textbf{plBART:} Wózki widłowe – zwiększamy wydajność i oszczędzamy paliwo

    \textbf{papuGaPT2:} Nowoczesny wózek widłowy – dlaczego jest tak ważny?

    \textbf{mT5-base:} Czym są nowoczesne technologie w wózkach widłowych?
 
    \textbf{plT5-large:} Automatyzacja wózków widłowych
\end{enumerate}

\section{Example predictions on Allegro Reviews \emph{body2lead}}
\begin{enumerate}
    \item \textbf{Gold Summary:} Każdy czuje respekt przed królową nauk. Matematyka spędza sen z powiek niejednemu uczniowi. Są jednak sposoby zachęcenia dziecka do nauki matematyki i sprawienia, że lepiej ją zrozumie i polubi.
    
    \textbf{plT5-base:} Matematyka to jedna z najprzyjemniejszych dziedzin nauki. Uczymy się ją w szkole i w domu. Nauka przez zabawę może być świetną rozrywką dla całej rodziny. Jakie gadżety mogą pomóc dziecku w nauce liczenia?
    
    \textbf{plBART:} Nauka matematyki nie musi być trudna. Wystarczy odrobina chęci i kilka akcesoriów, które pomogą dziecku w odkrywaniu tajemnic matematyki.
    
    \textbf{papuGaPT2:} Nauka poprzez zabawę to jeden z najlepszych sposobów na rozwijanie zdolności poznawczych i umiejętności logicznego myślenia. Podpowiadamy, co warto zabrać ze sobą do szkoły i na co zwrócić uwagę przy wyborze odpowiednich akcesoriów.
    
    \textbf{mT5-base:} Nauka matematyki to doskonały sposób na spędzenie wolnego czasu w gronie najbliższych. Dzieci chętnie sięgają po pomoce naukowe, które zachęcą do nauki przez zabawę i doświadczenie.
    
    \textbf{plT5-large:} Matematyka nie musi być nudna. Nauka liczenia może być wspaniałą zabawą, która wspomoże rozwój dziecka. Podpowiadamy, jak zachęcić dziecko do nauki matematyki.
    
    \item \textbf{Gold Summary:} Ciepłe, letnie dni aż się proszą o aktywną zabawę na dworze. Dlatego jeśli planujesz przyjęcie dla dziecka i jego przyjaciół, koniecznie postaw na garden party. Tego typu imprezy są nie tylko modne, ale także łatwe w przygotowaniu i posprzątaniu po nich. Jeśli nie masz pomysłu, jak je przygotować, podpowiadamy.

    \textbf{plT5-base:} Ozdabianie ogrodu to nie tylko świetna zabawa, ale także świetna rozrywka. Podpowiadamy, jak przygotować ogrodowe przyjęcie dla dziecka.

    \textbf{plBART:} Dzieci uwielbiają spędzać czas na świeżym powietrzu. To doskonała okazja do zorganizowania im urodzinowego przyjęcia, które zachwyci ich nie tylko wyglądem, ale także kreatywnością i zapewni wiele pozytywnych emocji. Podpowiadamy, jak przygotować party dla maluchów w cenie do 50 zł. 
    
    \textbf{papuGaPT2:} Organizujesz przyjęcie urodzinowe dla dziecka? Zastanawiasz się, co mu kupić, aby było mu jak najlepiej. Podpowiadamy, na co zwrócić uwagę, wybierając dekoracje do ogrodu. Sprawdź nasze propozycje.

    \textbf{mT5-base:} Ogrodowe przyjęcie to wyjątkowe wydarzenie w życiu każdego dziecka. To świetna okazja do spędzenia czasu na świeżym powietrzu. Warto więc zadbać o to, aby było przyjemne i kolorowe. Podpowiadamy, jakie dekoracje będą najlepsze dla twojego malucha. 
        
    \textbf{plT5-large} Ozdabianie ogrodu to nie tylko świetna zabawa, ale także świetna rozrywka. Podpowiadamy, jak przygotować ogrodowe przyjęcie dla dziecka.
\end{enumerate}

\section{Example predictions on PSC (whole)}\label{sec:example_psc_whole}
\begin{enumerate}
    \item \textbf{Gold Summary:} Ostatnia znaczna podwyżka stóp procentowych NBP oraz gwałtowne przyspieszenie inflacji wywołały krytykę poczynań Rady Polityki Pieniężnej. Rada chciałaby podjąć dyskusję z głosami krytycznymi. 
        nikt nie przewidywał, że w 1999 r. inflacja przewyższy cel inflacyjny NBP.Świadczy to dobitnie o tym, że głównymi czynnikami przyspieszającymi inflację były szoki podażowe na rynku żywności i paliw. Inny komentator stawia z kolei zarzut, że Rada niepotrzebnie zwlekała z decyzją o podwyżce stóp do listopada. Rada listopadową decyzję nie dlatego podjęła w listopadzie, i w takiej skali, że "zaspała" w październiku, ale dlatego, by podjąć ją w takiej właśnie skali w listopadzie. Dla każdego, kto wie, co to jest efektywność mechanizmu transmisji impulsów polityki pieniężnej do gospodarki, decyzja taka jest zrozumiała. Każdy ma prawo twierdzić, że dokonane podniesienie stóp NBP jest za duże. Pamiętać jednak musi, że jego twierdzenie sprowadza się do tezy: skala dokonanej podwyżki stóp procentowych doprowadzi do przestrzelenia celu inflacyjnego w dół. Czy rację mają ci, którzy krytykują teraz Radę, dowiemy się za rok. Przejdźmy do zarzutu zbyt dużej redukcji stóp procentowych 20 stycznia 1999 r. Otóż, redukcja rzeczywiście była zbyt duża bądź niepotrzebna. Problem polega jednak na tym, że o tym wiemy dopiero teraz. skala obniżki stóp procentowych dostosowana była do przewidywanego w 1999 r. przebiegu zjawisk makroekonomicznych. Było to działanie zgodne z zasadą forward looking. Czy Rada miała przesłanki do wcześniejszej zmiany swoich przewidywań? Bardzo mało i późno. o problemach ZUS dowiedzieliśmy się dopiero w lipcu. Gwałtowny wzrost cen żywności rozpoczął się od sierpnia. wzrosły ceny paliw. Z żadnych analiz nie wynikała taka skala wzrostu cen na rynku żywności i paliw. W tym kontekście stawianie Radzie zarzutu, że będąc odpowiedzialną za walkę z inflacją, stara się zrzucić winę na rząd i wydarzenia na rynku międzynarodowym, jest nierzetelne. Nie można winić straży pożarnej za lekkomyślność dziecka z zapałkami, jeśli okaże się sprawna w gaszeniu pożaru. nie można ignorować krótkookresowego wpływu na inflację czynników pozapieniężnych. nie da się na trwałe zmniejszyć inflacji bez budowy sprawnych i konkurencyjnych rynków paliw czy energii. 
        
        \textbf{mT5-small:} Ostatnia znaczna podwyżka stóp procentowych NBP oraz gwałtowne przyspieszenie inflacji wywołały ze strony niektórych ekonomistów, analityków i komentatorów krytykę poczynań Rady Polityki Pieniężnej. niestety, trudno podjąć merytoryczną dyskusję z głosami, szczególnie w sytuacji niepowodzeń.Rada listopadową decyzję podejmowała w całkowitym spokoju, bez żadnej nerwowości.Piędą zwracać uwagę na to, że polityka pieniężna była zbyt pożądanym, zważywszy rolę, jaką odgrywają. W tym kontekście stawianie Radzie zarzutów można było doprowadzić do końca września. 
        
        \textbf{plT5-small:} Ostatnia podwyżka stóp procentowych NBP oraz gwałtowne przyspieszenie inflacji wywołały krytykę poczynań Rady Polityki Pieniężnej. Rada listopadową decyzję podejmowała w całkowitym spokoju. Dla każdego, kto wie, co to jest efektywność mechanizmu transmisji impulsów polityki pieniężnej do gospodarki i jakie jest znaczenie oczekiwań w tym procesie, decyzja taka jest zrozumiała. redukcja stóp w styczniu rzeczywiście była zbyt duża bądź w ogóle niepotrzebna. 
        
        \textbf{papuGaPT2:} redukcja stóp w styczniu była zbyt duża bądź w ogóle niepotrzebna. skala krytycznych wniosków była dość trudna, stawiano wiele pytań, do których nie podejmowano ostatecznych decyzji.Rada miała przesłanki do wcześniejszej zmiany swoich przewidywań bardzo późno. informacje o wszystkich wydarzeniach dostępne były kilka tygodni po ich wystąpieniu. niską przewidywalność polityki gospodarczej rządu widać było gołym okiem. decyzja taka jest zrozumiała. redukcja stóp była koniecznym posunięciem, aby zbliżyć się do rynku. jednak Rada przyjmowała wówczas za wiarygodną zapowiedź zaostrzenia polityki fiskalnej oraz przewidywała powolny wzrost eksportu. starano się reanimować kryzys. pojawiła się propozycja, żeby Rada zaczęła mówić jednym głosem. Dziś grono to znacznie się powiększyło. z kolejnych wypowiedzi zaczęli wycofać wniosek o rewizję nadzwyczajną. Rada listopadową decyzję podejmowała w całkowitym spokoju, bez żadnej nerwowości. Dla każdego, kto wie, co to jest efektywność mechanizmu transmisji impulsów polityki do gospodarki i jakie jest znaczenie oczekiwań w tym procesie, decyzja o podwyżce stóp nie jest żadnym donosem. To oznacza duże prawdopodobieństwo żywiołowego narastania deficytu sektora publicznego. Ryzyko destabilizacji rynku jest wysokie. nie wolno zrezygnować z powrotu na przyzwoitą ścieżkę wzrostu ani z poprawienia równowagi zewnętrznej. Trzeba bardziej powściągliwie określić możliwości wycofania się inflacji z polityki monetarnej. jeśli spojrzymy na deprecjację parytetu, okazuje się, ze zestrzelenie nastąpiło nie z inicjatywy samego ROR-u, lecz z chęci ponownego ożywienia poprzez odpartyjnienie poprzez podjęcie działań, których w tej sytuacji jeszcze nie jesteśmy w stanie podjąć.

        \textbf{mT5-base:} Nie da się na trwałe zmniejszyć inflację bez budowy sprawnych i konkurencyjnych rynków paliw czy energii Szukanie winnego RYS. MARCIN CHUDZIK BoGUSŁAW GRABOWSKI Ostatnia znaczna podwyżka stóp procentowych NBP oraz gwałtowne przyspieszenie inflacji wywołały ze strony niektórych ekonomistów krytykę poczynań Rady Polityki Pieniężnej. Niestety, trudno podjąć merytoryczną dyskusję z głosami, szczególnie w sytuacji niepowodzeń w ograniczaniu inflacją. 
        
        \textbf{plBART:} Ostatnia znaczna podwyżka stóp procentowych NBP oraz gwałtowne przyspieszenie inflacji wywołały ze strony niektórych komentatorów krytykę poczynań Rady Polityki Pieniężnej. Rada listopadową decyzję podejmowała w całkowitym spokoju. Dla każdego, kto wie, co to jest efektywność mechanizmu transmisji impulsów polityki pieniężnej do gospodarki i jakie jest znaczenie oczekiwań w tym procesie, decyzja taka jest w pełni zrozumiała. redukcja stóp w styczniu rzeczywiście była zbyt duża bądź niepotrzebna. o tym wiemy dopiero teraz. stawianie Radzie zarzutu, że będąc odpowiedzialną za walkę z inflacją, stara się zrzucić winę na rząd, jest nierzetelne.
        
        \textbf{plT5-base:} Rząd Polityki Pieniężnej w poczuciu odpowiedzialności przywiązuje wielkie znaczenie do przejrzystości swoich działań od początku istnienia. Dlatego chciałaby podjąć dyskusję z głosami krytycznymi, szczególnie w sytuacji niepowodzeń w ograniczaniu inflacji. Znaczna część ostatnich krytycznych wypowiedzi wobec Rady sprowadzała się właściwie do trzech zarzutów: o to, że ostatnia podwyżka stóp procentowych była "spóźnioną, zbyt nerwową i przesadną" reakcją Rady, której "optymizm co do inflacji trwał zbyt długo", i która "jakby próbowała nadrobić zaległości w walce z inflacją".
        
        \item \textbf{Gold Summary:} Dwaj słynni bokserzy zawodowi Amerykanin Evander Holyfield i Brytyjczyk Lennox Lewis szykują się do rewanżowego pojedynku. Ich pierwszy pojedynek zakończył się, haniebnym według fachowców, remisem. Wygrany będzie królem boksu zawodowego.Historia boksu jest pełna ekscytujących rewanży, mają je również na swoim koncie Holyfield i Lewis.Obecny faworyt Lewis, jest aktualnym mistrzem świata WBC. Początkowy faworyt Holyfield, ma dwa pasy mistrzowskie WBA i IBF. Jeśli przegra Holyfield przegra tym samym Don King, najsłynniejszego promotora boksu zawodowego.Walkę transmituje Canal Plus.
        
        \textbf{mT5-small:} W najbliższą sobotę w Las Vegas dojdzie do kolejnego "rewanżu stulecia". Zmierzą się w nim dwaj najwybitniejsi obecnie bokserzy wagi ciężkiej, Amerykanin Evander Holyfield i Brytyjczyk Lennox Lewis. Osiem miesięcy temu, 13 marca, w nowojorskiej Madison Square Garden ich pierwsza walka zakończyła się kontrowersyjnym remisem. Stawką rewanżowego pojedynku będą pasy mistrzowskie trzech najważniejszych organizacji boksu zawodowego: WBA, WBC, IBF.
        
        \textbf{plT5-small:} W najbliższą sobotę w Las Vegas dojdzie do kolejnego "rewanżu stulecia". Zmierzą się w nim dwaj najwybitniejsi obecnie bokserzy wagi ciężkiej, Amerykanin Evander Holyfield i Brytyjczyk Lennox Lewis. Osiem miesięcy temu, 13 marca, w nowojorskiej Madison Square Garden ich pierwsza walka zakończyła się kontrowersyjnym remisem. Stawką rewanżowego pojedynku będą pasy mistrzowskie trzech najważniejszych organizacji boksu zawodowego: WBA, WBC, IBF. Ostatnim posiadaczem trzech pasów w wadze ciężkiej był Riddick Bowe. Od siedmiu lat żaden z pięściarzy nie może się uważać za króla tej najbardziej cenionej kategorii.
        
        \textbf{papuGaPT2:} W nocy z 13 na 14 listopada w Thomas and Mack Center w "Ready World Press" zmierzą się dwaj znakomici zawodnicy. Ich walkę będzie można oglądać na żywo w Canal Plus od godz. 3:25. Retransmisja w niedzielę będzie miała miejsce już raz trzeci w historii walki, która miała być największym wydarzeniem ostatnich lat. Transmitowany przez Jamesa "Bustera" pojedynek Sonny'ego Listona z Bobem Fitzsimmonsem w 1899 r. zakończył się dyskwalifikacją jednego z zawodników. Walka okazała się skandalem. Trzy niezależne postępowania wyjaśniające, które prowadzono po tej walce, niewiele dały. Nikomu nie udowodniono winy. \textbackslash nBoks jest tak stary jak świat. Już Homer opiewał walki na gołe pięści. Wielkim zwolennikiem walk bokserów był rzymski cesarz Kaligula. W 1350 roku, gdy naprzeciw siebie stanęli Jeffries i O'Donnnel, dwóch zawodników zadało przeciwnikowi dwa razy więcej ciosów, niż spalonych. Dwudziestotrzyletni wówczas Buster trafił prawym prostym. Teraz, na dziesięć metrów, już w pierwszej rundzie Floyda Pattersona spotkał się z Joem Frazierem, Muhammadem Alim, Tunneyem i Michaelem Moorerem. O 3.00 rano, kiedy stoczyli już swoje ostatnie rugi, przyszedł czas na jeszcze jedną walkę z "Milionerem", "Drakula" z Jackiem Dempseyem.

        \textbf{mT5-base:} W najbliższą sobotę w Las Vegas dojdzie do kolejnego "rewanżu stulecia". Zmierzą się w nim dwaj najwybitniejsi obecnie bokserzy wagi ciężkiej, Amerykanin Evander Holyfield i Brytyjczyk Lennox Lewis. Osiem miesięcy temu, 13 marca, w nowojorskiej Madison Square Garden ich pierwsza walka zakończyła się kontrowersyjnym remisem. Stawką rewanżowego pojedynku będą pasy mistrzowskie trzech najważniejszych organizacji boksu zawodowego: WBA, WBC, IBF. 
        
        \textbf{plBART:} W najbliższą sobotę w Las Vegas dojdzie do kolejnego "rewanżu stulecia". Zmierzą się w nim dwaj najwybitniejsi obecnie bokserzy wagi ciężkiej, Amerykanin Evander Holyfield i Brytyjczyk Lennox Lewis. Osiem miesięcy temu, 13 marca, w nowojorskiej Madison Square Garden ich pierwsza walka zakończyła się kontrowersyjnym remisem. Stawką rewanżowego pojedynku będą pasy mistrzowskie trzech najważniejszych organizacji boksu zawodowego: WBA, WBC, IBF. Ostatnim posiadaczem trzech pasów w wadze ciężkiej był Riddick Bowe. Od siedmiu lat żaden z pięściarzy nie może się uważać za króla tej najbardziej cenionej kategorii.
        
        \textbf{plT5-base:} W najbliższą sobotę w Las Vegas dojdzie do kolejnego "rewanżu stulecia". Zmierzą się w nim dwaj najwybitniejsi obecnie bokserzy wagi ciężkiej, Amerykanin Evander Holyfield i Brytyjczyk Lennox Lewis. Osiem miesięcy temu, 13 marca, w nowojorskiej Madison Square Garden ich pierwsza walka zakończyła się kontrowersyjnym remisem. Stawką rewanżowego pojedynku będą pasy mistrzowskie trzech najważniejszych organizacji boksu zawodowego: WBA, WBC, IBF. Ostatnim posiadaczem trzech pasów w wadze ciężkiej był Riddick Bowe. Od siedmiu lat żaden z pięściarzy nie może się uważać za króla tej najbardziej cenionej kategorii.
\end{enumerate}

\section{Example predictions on PSC (extract)}\label{sec:example_psc_extract}
\begin{enumerate}
    \item \textbf{Gold Summary:} Politycy z pierwszych stron gazet szykują się do objęcia stanowisk prezydentów dużych miast. Dla nich gotowi są nawet zrezygnować z mandatów poselskich.\textbackslash nW najbliższych tygodniach rozstrzygnie się, czy w jesiennych wyborach samorządowych prezydenci miast wybierani będą bezpośrednio przez mieszkańców. W Sejmie trwają prace nad projektem ustawy o bezpośrednim wyborze wójtów, burmistrzów i prezydentów. Ustawa ma poparcie SLD i PO. Jako kandydatów SLD na stanowisko w stołecznym ratuszu wymienia się Ryszarda Kalisza i Danutę Waniek. "Nie wyklucza" kandydowania w Warszawie Lech Kaczyński z Prawa i Sprawiedliwości. Trzecim liczącym się kandydatem może być Andrzej Olechowski, lider Platformy Obywatelskiej. Nie stanie w szranki natomiast były prezydent Paweł Piskorski. 
    
    \textbf{mT5-small:} Dla nich gotowi są nawet zrezygnować z mandatów poselskich. W najbliższych tygodniach rozstrzygnie się, czy w jesiennych wyborach samorządowych prezydenci miast wybierani będą bezpośrednio przez mieszkańców. ako kandydatów SLD na stanowisko w stołecznym ratuszu wymienia się Ryszarda Kalisza i Danutę Waniek.W partii krąży opinia, że Kalisz sam się zgłosił.
    
    \textbf{plT5-small:} Politycy szykują się do objęcia stanowisk prezydentów dużych miast. Dla nich gotowi są nawet zrezygnować z mandatów poselskich. W najbliższych tygodniach rozstrzygnie się, czy w jesiennych wyborach samorządowych prezydenci miast wybierani będą bezpośrednio przez mieszkańców.
    
    \textbf{papuGaPT2:} Partia ma poważne problemy z przywództwem. Nie ma mocnego nazwiska. Na razie próbuje się rozwiązywać kolejne oświęcimskie kryzysy w drodze wykorzystywania kruczków prawnych. Niedawno wyszły na jaw kontakty radnego z półświatkiem.W Szczecinie najpoważniejszym z kandydatów jest Edmund Runowicz, znany wszystkim z Gliwic lider ZChN. Z racji wpływów jakie ma partyjniactwo w mieście, chcą go popierać.Próba stworzenia drugiego centrum politycznego nie powiodła się. SLD zrezygnował z pomysłu utworzenia szerokiej koalicji ugrupowań posierpniowych. Przy budżecie państwa 181,6 mld działalność biura wyborczego kosztuje 0,041 proc. państwowych wydatków.Tymczasem w ramach walki z bezrobociem szuka ratunku w Urzędzie Miasta.

    \textbf{mT5-base:} W najbliższych tygodniach rozstrzygnie się, czy w jesiennych wyborach samorządowych prezydenci miast wybierani będą bezpośrednio przez mieszkańców. Tymczasem konkurenci sięgają po najlepsze kadry. Sojusz ma kłopot, bo nie ma drugiego tak dobrego kandydata. Jak nie Borowski, to kto? Jako kandydatów SLD na stanowisko w stołecznym ratuszu wymienia się Ryszarda Kalisza i Danutę Waniek. Te dwie osoby mają w stolicy duże poparcie.
    
    \textbf{plBART:} Politycy szykują się do objęcia stanowisk prezydentów dużych miast. Dla nich gotowi są nawet zrezygnować z mandatów poselskich.W najbliższych tygodniach rozstrzygnie się, czy w jesiennych wyborach samorządowych prezydenci miast wybierani będą bezpośrednio przez mieszkańców. Jako kandydatów SLD na stanowisko w stołecznym ratuszu wymienia się Ryszarda Kalisza i Danutę Waniek. "Nie wyklucza" kandydowania w Warszawie Lech Kaczyński z Prawa i Sprawiedliwości. Trzecim liczącym się kandydatem może być Andrzej Olechowski, lider Platformy Obywatelskiej.
    
    \textbf{plT5-base:} Czy w jesiennych wyborach samorządowych prezydenci miast wybierani będą bezpośrednio przez mieszkańców. W Sejmie trwają prace nad projektem ustawy o bezpośrednim wyborze wójtów, burmistrzów i prezydentów. Ustawa ma poparcie głównych sił politycznych, jak SLD i PO, jest więc wielce prawdopodobne, iż zostanie uchwalona. Nic dziwnego, że już dzisiaj partie zastanawiają się, kim obsadzić najwyższe stanowiska w miastach. Dla niektórych polityków odejście do samorządu oznaczałoby konieczność złożenia mandatu posła, ale nie zraża ich to.
    
    \item \textbf{Gold Summary:} Ludzie całkowicie sparaliżowani będą mogli samą myślą sterować komputerem, by porozumieć się z otoczeniem. Niezwykle czułe detektory, odczytujące wybrane fale mózgowe, testowane są w laboratoriach USA, Japonii i Europy. Do łączności z urządzeniami elektronicznymi można wykorzystać jakąkolwiek aktywność układu nerwowego - sygnały elektryczne wytwarzane ruchami gałek ocznych lub impulsy nerwów mięśni.
    
    \textbf{mT5-small:} Uniwersytetu Emory w Atlancie wypróbowali u dwóch pacjentów system pozwalający sterować kursorem na ekranie monitora. Sterowanie komputerem samą myślą stało się możliwe, gdy uczeni lepiej opanowali odczytywanie i werbalizację sygnałów. Komunikowanie się przy użyciu takiego systemu jest dość żmudne, ale wyłącznie dla ludzi zdrowych. Dla osób całkowicie sparaliżowanych, nie mogących wykonywać żadnych ruchów, nawet mięśniami twarzy, jest jedynym sposobem porozumiewania.
    
    \textbf{plT5-small:} Ludzie całkowicie sparaliżowani będą mogli samą myślą sterować komputerem, by porozumieć się z otoczeniem. Niezwykle czułe detektory, odczytujące wybrane fale mózgowe, testowane są w laboratoriach USA, Japonii i Europy. Do łączności z urządzeniami elektronicznymi można wykorzystać jakąkolwiek aktywność układu nerwowego - sygnały elektryczne wytwarzane ruchami gałek ocznych lub impulsy nerwów mięśni. Jeszcze niedawno wszelkie rozważania na ten temat były domeną wyłącznie autorów powieści fantastycznonaukowych. Obecnie naukowcy w kilkunastu ośrodkach prześcigają się w konstruowaniu urządzeń coraz sprawniej wykrywających komunikaty myślowe.
    
    \textbf{papuGaPT2:} dzisiaj coraz więcej teoretyków na temat tego, co może się stać, jak będzie możliwe, za kilka lat. Trudno wyobrazić sobie wielką liczbę skomplikowaną, kręcących się po krzywej Kenii i Epsilon. Gdyby jednak wziąć pod uwagę normalny, a nie ten podkolorowany obraz powyborczej rzeczywistości, to rzeczywiście wyglądałby nędzny. Jeszcze niedawno takie pytanie byłoby niedorzeczne. Dziś naukowcy nie są już takimi sceptykami. \textbackslash nNiektórzy jeszcze przed przystąpieniem Polski do UE zastanawiali się, czy w ogóle wchodzić w ten projekt. Nie wszyscy są entuzjastami nowego tworu, który stał się również przedmiotem marzeń i zazdrości milionów. A Polska? Jak wynika z sondaży, 77 proc. Amerykanów gotowych jest poprzeć likwidację lub gruntowną zmianę europejskich norm. Przeciwnicy wejścia do Unii Europejskiej obawiają się takiego myślenia, bo będzie on trwał tylko trzy i pół roku. Natomiast w rzeczywistości jest to proces bardzo długo.
    
    \textbf{mT5-base:} Jeszcze niedawno wszelkie rozważania na ten temat były domeną wyłącznie autorów powieści fantastycznonaukowych. Obecnie uczeni w kilkunastu ośrodkach prześcigają się w konstruowaniu urządzeń coraz sprawniej wykrywających komunikaty myślowe. Niezwykle czułe detektory, odczytujące wybrane fale mózgowe, testowane są w laboratoriach USA, Japonii i Europy. Kilka takich rozwiązań jest testowanych w USA.
        
    \textbf{plBART:} Ludzie całkowicie sparaliżowani będą mogli samą myślą sterować komputerem, by porozumieć się z otoczeniem. Niezwykle czułe detektory, odczytujące wybrane fale mózgowe, testowane są w laboratoriach USA, Japonii i Europy. Do łączności z urządzeniami elektronicznymi można wykorzystać jakąkolwiek aktywność układu nerwowego. Urządzenie ma jednak tę wadę, że wymaga użycia wszczepów, grożących powstaniem infekcji i uszkodzeniem mózgu.
    
    \textbf{plT5-base:} ZBIGNIEW WOJTASIŃSKI Ludzie całkowicie sparaliżowani będą mogli samą myślą sterować komputerem, by porozumieć się z otoczeniem. Niezwykle czułe detektory, odczytujące wybrane fale mózgowe, testowane są w laboratoriach USA, Japonii i Europy. Do łączności z urządzeniami elektronicznymi można wykorzystać jakąkolwiek aktywność układu nerwowego - sygnały elektryczne wytwarzane ruchami gałek ocznych lub impulsy nerwów mięśni.
\end{enumerate}

\section{Example predictions on PSC (abstract)}\label{sec:example_psc_abstract}
\begin{enumerate}
    \item \textbf{Gold Summary:} Anachroniczna jest koncepcja, u nas bynajmniej nierzadko wyznawana, że swoboda umów wyklucza ideę ochrony konsumenta.Przed kilkoma laty niemiecki Trybunał Konstytucyjny uznał, że nawet osoba pełnoletnia, samodzielna, nie poddana żadnemu przymusowi wymaga ochrony ręcząc za kredyt bankowy oraz że konieczna jest szczegółowa, wyczerpująca, niedwuznaczna, jasna informacja, ze wskazaniem na kwestie najbardziej niebezpieczne dla poręczyciela. Lepsza informacja dla konsumenta łagodzi bowiem nierówność pozycji rynkowej. W konsekwencji tych orzeczeń zmieniła się praktyka powszechnych sądów w Niemczech. Trzy kwestie zasługują tu na uwagę. Po pierwsze - umowy kredytowe i sytuacja poręczyciela doczekały się w Niemczech oceny TK, dokonywanej z konstytucyjnego punktu widzenia. Po drugie - TK za remedium na strukturalne zachwianie równowagi umownej uznał zwiększenie obowiązków informacyjnych kontrahenta konsumenta. Po trzecie wreszcie - opisywana sytuacja jest kolejnym przykładem tego, jak dalece anachroniczna jest koncepcja (u nas bynajmniej nierzadko wyznawana), że swoboda umów wyklucza ideę ochrony konsumenta.Z konstytucji nie można wyczytać, jakimi środkami i na jakim poziomie ma się chronić konsumenta, ale to nie jest jedyny możliwy sposób "użycia konstytucji" do takich celów. Ustawa zasadnicza da się użyć jako norma rozstrzygająca na wypadek kilku możliwych interpretacji jakiegoś przepisu: "prokonsumenckiej" (w zakresie wymienionych w konstytucji, szczególnie chronionych praw konsumenta) i "antykonsumenckiej" lub choćby "konsumencko neutralnej". Albo na wypadek interpretacji norm blankietowych, klauzul generalnych czy zwrotów niedookreślonych, które sąd musi odkodować, nadać im konkretną treść.Drugą ważną kwestią jest znaczenie informacji jako oręża konsumenta. W europejskim prawie wspólnotowym kamieniem węgielnym ochrony konsumenta jest informacja. Uważa się, że konsument wymaga ochrony, ponieważ jest źle poinformowany i na skutek tego nie może w prawdziwie wolny i nieskrępowany sposób decydować o swym "udziale na rynku". Stąd się biorą niezwykle rozbudowane w dyrektywach i ich implementacjach wewnątrzkrajowych przepisy mówiące, o czym, kiedy i jak trzeba konsumenta informować. Nie tylko jednak sama obrona przez informację jest cechą charakterystyczną europejskiego prawa konsumenckiego. Występuje tu jeszcze wskazanie minimalnego poziomu treści umowy. Oznacza to, że poszczególne państwa, implementując dyrektywy konsumenckie we własnym porządku prawnym, nie mogą zejść niżej wskazanego tam poziomu ochrony. Ale tutaj tkwi pułapka: przyznanie w prawie krajowym zbyt wysokiego poziomu ochrony, może być uznane za rzeczywistą czy ukrytą praktykę dyskryminacyjną, ograniczającą dostęp towarów i usług na rynek wspólny. Nie tylko informacja i nie tylko określenie minimalnego poziomu uprawnień zagwarantowanych konsumentowi w dotyczącym danej umowy prawie wspólnotowym czy wewnętrznym są cechami charakterystycznymi europejskiego prawa konsumenckiego. Otóż cechą charakterystyczną współczesnego prawa umów jest to, że konsumentowi przysługuje prawo "wycofania się z transakcji". W praktyce i u nas ta nowa technika ochronna zaczyna się pojawiać: zna ją polskie prawo ubezpieczeniowe. Co charakterystyczne: tego rodzaju możliwość wycofania się z transakcji służy (w prawie europejskim) jedynie konsumentowi, nie jego kontrahentowi. I znów jest to jeden ze środków, który przyznaje się tylko stronie instytucjonalnie słabszej, używany, by zrównoważyć siłę rynkową konsumenta wobec profesjonalisty. 
    
    \textbf{mT5-small:} Pod koniec laty dwa orzeczenia niemieckiego Trybunału Konstytucyjnego zbulwersowały światek prawniczy. Trybunał uznał, że anachroniczna jest koncepcja, u nas bynajmniej nierzadko wyznawana, jako swoboda umów wyklucza ideę ochrony konsumenta. Po trzecie - umowy kredytowe i sytuacja poręczyciela doczekały się w Niemczech oceny TK, dokonywanej z konstytucyjnym punktu widzenia.
    
    \textbf{plT5-small:} Przed kilkoma laty dwa orzeczenia niemieckiego Trybunału Konstytucyjnego zbulwersowały świat prawniczy. Trybunał uznał, że nawet osoba pełnoletnia, samodzielna i niepoddana żadnemu przymusowi wymaga ochrony ręcząc za kredyt bankowy. Drugą istotną kwestią europejskiego prawa konsumenckiego jest wskazanie minimalnej wiedzy i doświadczenia między poręczycielem i bankiem. Innym przykładem środka ochrony konsumenta jest niemiecki Trybunał Konstytucyjny. Moim zdaniem, konsumentowi przysługuje prawo wycofania się z transakcji bez negatywnych konsekwencji.
    
    \textbf{papuGaPT2:} W porównaniu z prawem polskim ustawodawstwo i orzecznictwo krajów europejskich znacznie bardziej chroni prawa konsumentów, co w konsekwencji doprowadzić może do unikania przez banki i inne instytucje finansowe prokonsumenckie wszelkiej informacji o towarze i bezpiecznym sposobie korzystania z niego. Europejskie dyrektywy konsumenckie wskazują minimalny poziom treści umowy: są to przepisy zakazujące reklamy pewnych produktów czy usług (lekarstwa, alkohole, wolne zawody), używania zrozumiałego w określonych mediach (radio, tv), reklamy kierowanej do pewnych osób (dzieci) czy też zniósł dobrze ochronę przed wprowadzającymi w błąd informacjami na etykietach. Wreszcie cechą charakterystyczną współczesnego prawa umów jest to, iż konsument przez krótki czas (siedem do dziesięciu dni) ma prawo, bez żadnych negatywnych konsekwencji, zrezygnować z transakcji. Środek ten przysługuje tylko konsumentowi jako stronie instytucjonalnie słabszej; jego kontrahenta obowiązuje ogólna zasada mówiąca o "umowy powinny zostać dotrzymywane". Zasada ta wyraża się w łacińskiej formule "in dubio contra proferentem" i jest ważnym instrumentem prawa cywilnego, który służy nie tylko ochronie konsumenta, ale także jako środek prawnokarnego (art. 76), który ma zrównoważyć siłę przyciągania i działać na rzecz zrekompensowania braków jego wiedzy. Stąd się biorą rozbudowane przepisy mówiące, o czym, kiedy i jak należy informować konsumenta o wszelkich aspektach umowy kredytowej, a także o możliwych niebezpieczeństwach i ryzykach, które mogą być nieświadomym wyrazem orientacji konsumenta w transakcjach rynkowych.

    \textbf{mT5-base:} Rę swoboda umów wyklucza ideę ochrony konsumenta.
    
    \textbf{plBART:} Przed kilkoma laty dwa orzeczenia niemieckiego Trybunału Konstytucyjnego zbulwersowały świat prawniczy. Trybunał uznał, że nawet osoba pełnoletnia, samodzielna, nie poddana żadnemu przymusowi wymaga ochrony ręcząc za kredyt bankowy, bo nierównowaga wiedzy i doświadczenia między poręczycielem i bankiem jest zbyt wielka i tolerowanie jej zagraża jednej z konstytucyjnych zasad niemieckiego porządku prawnego: wyrażonej w art. 2 konstytucji autonomii woli. W konsekwencji orzeczeń TK zmieniła się praktyka niemieckich sądów, które umowy kredytowe doczekały się oceny TK, dokonywanej z konstytucyjnego punktu widzenia. Po drugie, TK uznał za remedium na strukturalne zachwianie równowagi umownej. Opisywana sytuacja jest kolejnym przykładem tego, jak dalece anachroniczna jest koncepcja (u nas bynajmniej nierzadko wyznawana), że swoboda umów wyklucza ideę ochrony konsumenta.
    
    \textbf{plT5-base:} W.
    
    \item \textbf{Gold Summary:} Praca Huberta Orłowskiego "Polnische Wirtschaft. Nowoczesny niemiecki dyskurs o Polsce" poświęcona jest stereotypom dotyczącym stosunków polsko-niemieckich, w tym jednemu z najsilniejszych - "polnische Wirtschaft". Autorstwo tego stereotypu przypisuje się Georgowi Fosterowi, uczonemu i podróżnikowi, który w swoich listach pisanych pod koniec XVIII w. terminem tym określał kombinację bałaganu, nieudolności, brudu i lenistwa. W stereotyp myślenia Niemców o Polsce wpisywały się wydarzenia historyczne. Zrywy niepodległościowe Polaków od powstania listopadowego do "Solidarności" traktowane były często jako przejaw polskiej skłonności do anarchii. We współczesnych Niemczech stereotyp "polnische Wirtschaft" jest wciąż żywy. Podtrzymują go zarówno media, jak i w literatura piękna.
    
    \textbf{mT5-small:} Nakładem wydawnictwa Huberta Orłowskiego poświęcona jest stereotypom dotyczącym stosunków polsko-niemieckich, ze szczególnym uwzględnieniem najsilniejszych z nich: „polnische Wirtschaft”. Autorstwo tego stereotypu przypisuje się niejakiemu Georgowi Fosterowi, uczonemu i podróżnikowi z Żuław Wiślanych. "Polnische Wirtschaft" oznaczającemu kombinację tych cech. Wśród warstw bogatszych widziano rozrzutność, równocześnie jednak także brak higieny, zacofanie, ignorancja.
    
    \textbf{plT5-small:} W książce profesora Huberta Orłowskiego poświęcona jest stereotypom dotyczącym stosunków polsko-niemieckich, ze szczególnym uwzględnieniem najsilniejszych z nich: "polnische Wirtschaft" oznaczającemu kombinację bałaganu, nieudolności, brudu i złą gospodarkę.
    
    \textbf{papuGaPT2:} Od lat 40. literatura podtrzymywała mit szlachetnego Polaka, o którym zapomina młodsze pokolenie. Stereotyp ten nigdy by może nie zaistniał, gdyby nie głębokie różnice dzielące polskie cnoty szlacheckie od niemieckich cnót mieszczańskich, pogłębione różnicami między katolicyzmem i protestantyzmem. Po klęsce powstania listopadowego, po upadku muru berlińskiego okazało się, że ustrój ten nie był w stanie zdobyć uznania i poparcia ludu, dlatego też w książkach, operetkach i filmach pełno jest dowcipów o polskiej gospodarce i polskim sejmikowaniu. Profesor Hubert Orłowski jest współautorem tzw. ustawy teczkowej, ustanawiającej Instytut Pamięci Narodowej, oraz ustawy o ochronie informacji niejawnych, określającej m.in. mianem "pacta servanda sunt" na której spoczywa cała odpowiedzialność.
    
    \textbf{mT5-base:} W niemieckich zbiorów przysłów o Polsce, ukazała się dwa lata temu w Wiesbaden nakładem wydawnictwa Otto Harrasowitza na polski.
    
    \textbf{plBART:} Praca profesora Huberta Orłowskiego "Polnische Wirtschaft. Nowoczesny niemiecki dyskurs o Polsce" poświęcona jest stereotypom dotyczącym stosunków polsko-niemieckich, ze szczególnym uwzględnieniem najsilniejszych z nich: „polnischer Wirthschaft”. Termin ten określa "nieporządek i złą gospodarkę, niedbalstwo i bałagan". Autorstwo tego, mającego długą historię i niesłychaną żywotność, stereotypu przypisuje się niejakiemu Georgowi Fosterowi, uczonemu i podróżnikowi z Żuław Wiślanych. Naprawdę jednak formuła była już gotowa i nieźle znana wcześniej, choć w myśleniu Niemców (i nie tylko ich) zadomowiła się dopiero ok. 1830 r. po edycji listów Fostera. W liście do swojego wierzyciela pisał z Wilna w 1784 roku: "O nieopisanym brudzie, lenistwie, opilstwie i nieudolności całej służby..., o niezdarności rzemieślników, ich niesłychanie kiepskiej robocie, wreszcie o zadowoleniu Polaków [Polaken] z własnego bagienka, a także ich przywiązaniu do rodzinnych zwyczajów nie chcę pisać już nic więcej, aby nie przedłużać tego listu".
    
    \textbf{plT5-base:} W. 
\end{enumerate}
\end{document}